\pdfoutput=1

\documentclass[11pt]{article}

\usepackage{acl}

\usepackage{times}
\usepackage{latexsym}

\usepackage[T1]{fontenc}

\usepackage[utf8]{inputenc}

\usepackage{microtype}

\usepackage{inconsolata}

\usepackage{times}
\usepackage{latexsym}
\usepackage{enumitem}
\usepackage{multirow}
\usepackage{array}
\usepackage{amsfonts}
\usepackage{subcaption}
\usepackage{makecell}
\usepackage{soul}
\usepackage{xcolor,colortbl}
\usepackage{xspace}
\usepackage{textcomp}
\usepackage{stfloats}
\usepackage{url}
\usepackage{verbatim}
\usepackage{graphicx}
\usepackage{algorithm}
\usepackage{algorithmic}
\usepackage{amssymb}
\usepackage{helvet}  
\usepackage{courier}  
\usepackage{newfloat}
\usepackage{listings}
\usepackage{multirow}
\usepackage{amsmath}
\usepackage{booktabs}
\usepackage{pifont}
\usepackage{color}
\usepackage{bbding}
\usepackage[export]{adjustbox}

\newcommand{{\model}}{InstructGraph}

\title{\protect\includegraphics[scale=.04, valign=c]{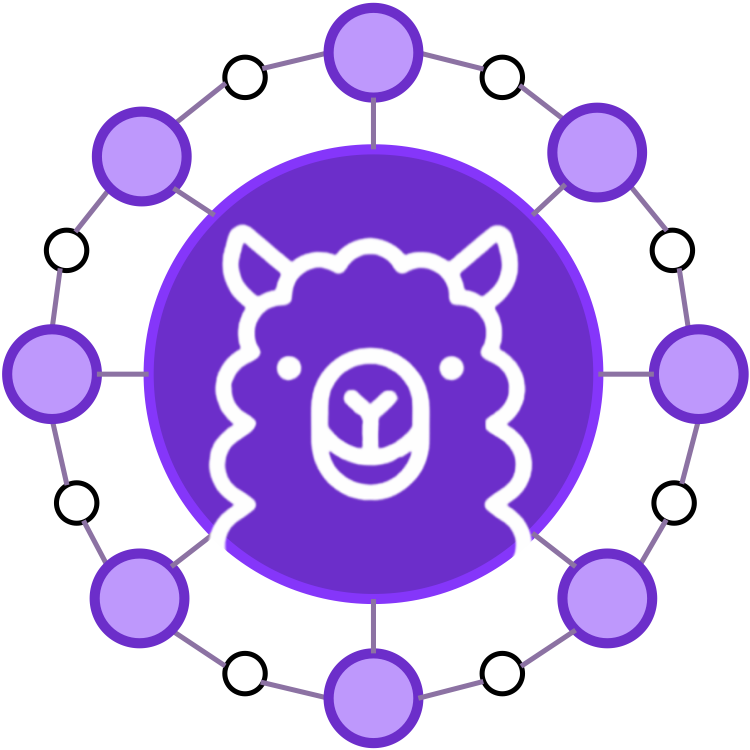} InstructGraph: Boosting Large Language Models via Graph-centric Instruction Tuning and Preference Alignment}

\author{Jianing Wang$^{1,2}$\thanks{\  Work done during visiting at UC San Diego.}, Junda Wu$^{2}$, Yupeng Hou$^{2}$, Yao Liu$^{1}$\thanks{\ Corresponding Author.}, Ming Gao$^{1}$, Julian McAuley$^{2}$\\
  $^1$ East China Normal University, Shanghai, China \\
  $^2$ University of California San Diego, La Jolla, USA\\
  \texttt{lygwjn@gmail.com,} \texttt{\{juw069, yphou\}@ucsd.edu} \\
  \texttt{liuyao@cc.ecnu.edu.cn,}
  \texttt{mgao@dase.ecnu.edu.cn,}
  \texttt{jmcauley@ucsd.edu}
}

\begin{document}
\maketitle
\begin{abstract}

Do current large language models (LLMs) better solve graph reasoning and generation tasks with parameter updates?
In this paper, we propose \textbf{InstructGraph},
a framework that empowers LLMs with the abilities of graph reasoning and generation by instruction tuning and preference alignment.
Specifically, we first propose a structured format verbalizer to unify all graph data into a universal code-like format, 
which can simply represent the graph without any external graph-specific encoders.
Furthermore, a graph instruction tuning stage is introduced to guide LLMs in solving graph reasoning and generation tasks.
Finally, 
we identify potential hallucination problems in graph tasks and sample negative instances for preference alignment, the target of which is to enhance the output's reliability of the model.   
Extensive experiments across multiple 
graph-centric tasks exhibit that InstructGraph can achieve the best performance and outperform GPT-4 and LLaMA2 by more than 13\% and 38\%, respectively~\footnote{We have released the resource code in \url{https://github.com/wjn1996/InstructGraph}.}.
\end{abstract}

\section{Introduction}


Currently, large language models (LLMs) have succeeded in reasoning on textual data~\cite{Brown2020Language, OpenAIGPT2023, Touvron2023Llama2, Zhao2023A}.
However, there also exists rich information in graph data, that is difficult to represent using plain text~\cite{Jin2023Large}, 
such as knowledge graphs~\cite{Schneider2022A}, symbolic graphs~\cite{Saba2023Stochastic}, social networks~\cite{Wang2023Rethinking}, and implicit mind graphs~\cite{Besta2023Graph}. 

To endow LLMs with the ability to 
solve graph tasks, 
a series of works focus on designing the interface (e.g., prompt engineering) of LLMs on graph data to make them understand the semantics without parameter optimization~\cite{Ye2023Natural, Han2023PiVe, Zhang2023LLM4DyG, Zhang2023Graph, Kim2023KGGPT, Jiang2023StructGPT, Wang2023Boosting, Luo2023Reasoning}, or injecting the graph embeddings into the partial parameters of LLMs through graph neural networks (GNNs)~\cite{Zhang2022GreaseLM, Chai2023GraphLLM, Tang2023GraphGPT, Perozzi2024Let}.
Despite significant progress, we explore these two challenges:
1) There still exists a semantic gap between graph and text, 
which may impede the LLM in graph reasoning and generation.
2) LLMs tend to generate hallucinations which may be caused by fabricated erroneous inputs or lack of pertinent knowledge. It can be viewed as the graph hallucination problem.

\begin{figure*}
    \centering
	\includegraphics[width=\linewidth]{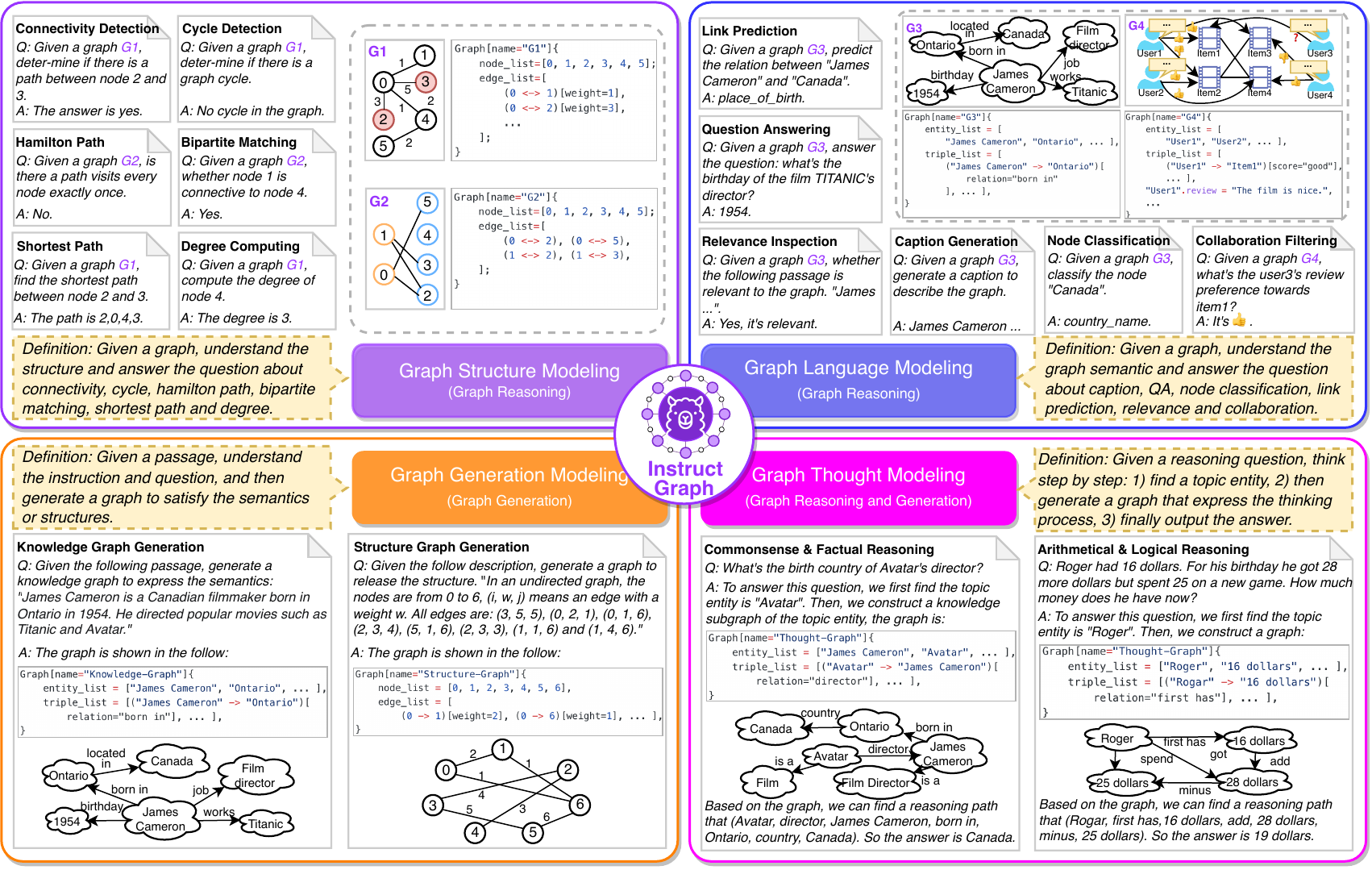}
	\caption{Four groups of graph-centric reasoning and generation tasks.}
 \label{fig:instructgraph_all_task}
 \vspace{-0.5em}
\end{figure*}


To overcome these challenges, we present a framework named \textbf{InstructGraph} that boosts LLMs by instruction tuning and preference alignment.
A straightforward approach to solve the first challenge is to use a graph description~\cite{Ye2023Natural} or graph embeddings~\cite{Chai2023GraphLLM},
However, these methods require a large number of
manual templates to describe the graph.
Representing a large or complex graph via
embeddings may cause information loss. 
In addition, the responses generated by the LLM with these methods are difficult to parse into actual graphs~\cite{Jin2023Large, Zhao2023A}.
Current investigations have demonstrated that LLMs have a great ability for code understanding and generation~\cite{Gao2023What, Ma2023At, Wong2023Natural, Yang2024If}. 
Inspired by them, we can unify graph data into a code-like universal format to enhance the LLM's understanding and generation performance on graph tasks.
As shown in Figure~\ref{fig:instructgraph_all_task}, each graph can be converted into a code with basic variables, such as \texttt{node\_list} (or \texttt{entity\_list}), \texttt{edge\_list} (or \texttt{triple\_list}) and optional properties.
To this end, a graph instruction tuning stage is introduced to train the LLM on these formulated data.

In addition, previous works
have found that 
LLMs generate responses with hallucination when following the instructions,
typically referring to fabricated erroneous inputs or lack of intrinsic knowledge~\cite{Dziri2022On, Zhang2023Siren, Ji2023Survey}.
For example, the LLM may derive a wrong answer when being questioned on a graph that lacks key information, or the LLM may generate a graph with incorrect facts, conflicting, or missing information.
However, how to reduce this effect in graph reasoning and generation is still under-explored.
Hence, we introduce the graph preference alignment to alleviate the hallucination problem in the LLM's reasoning and generation. 
Specifically, we follow the direct preference optimization (DPO) algorithm~\cite{Rafailov2023Direct} to optimize the LLM to make better preferences.
To automatically sample the negative instances in DPO, we explore 
various scenarios,
such as \emph{unfactual graph}, \emph{conflict graph} and \emph{missing graph}.
, to simulate the graph hallucination problem. 

To evaluate the effectiveness of our framework, we perform extensive experiments on multiple graph reasoning and generation tasks.
Results 
reveal
that the proposed InstructGraph achieves the best performance on both graph-centric instruction and preference tasks and outperforms the GPT-4~\cite{OpenAI2023GPT4} and LLaMA2~\cite{Touvron2023Llama2} by more than 13\% and 38\%, respectively.

\section{Methodology}

\begin{figure*}[t]
\centering
\includegraphics[width=\linewidth]{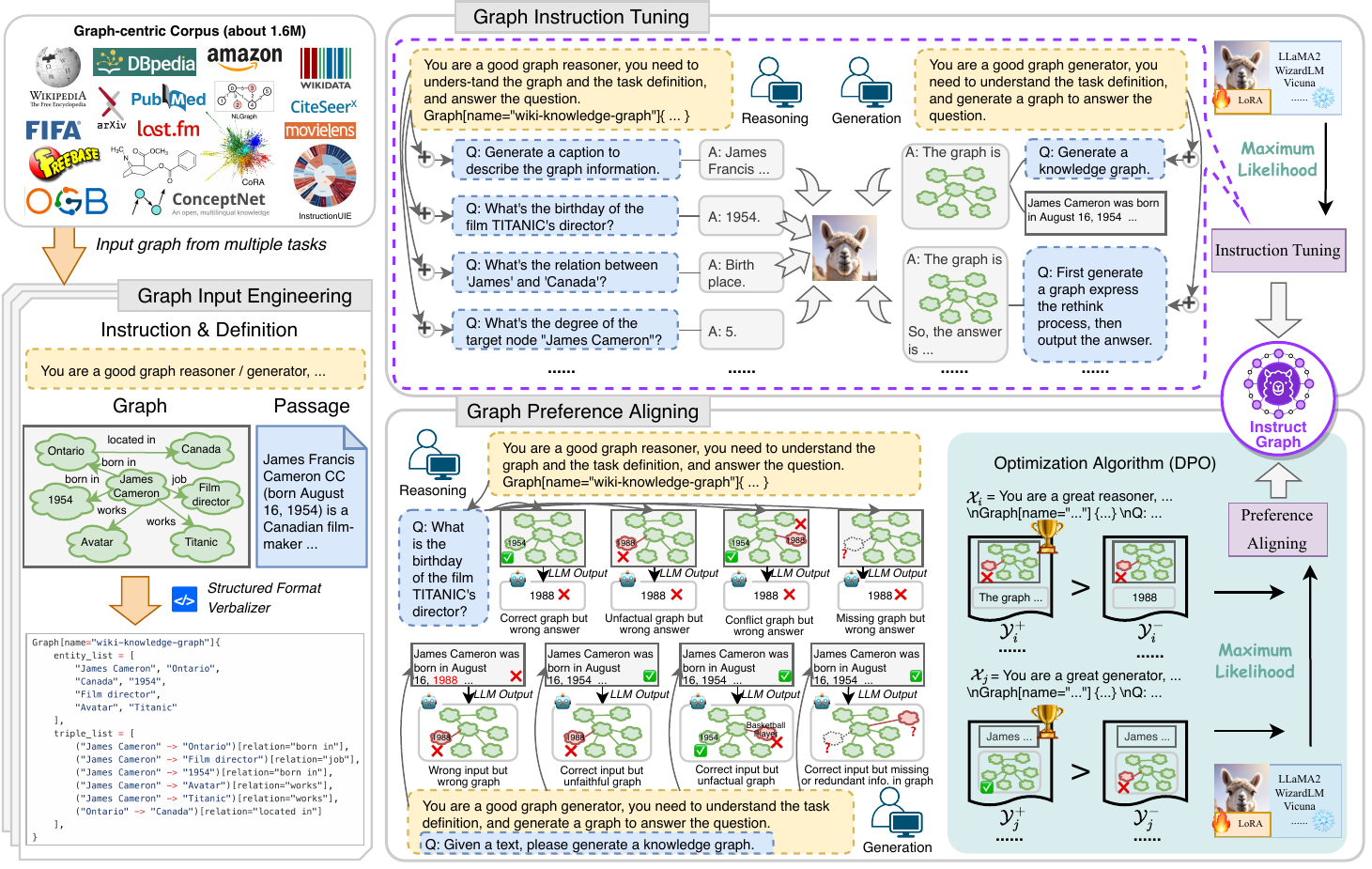}
\caption{The InstructGraph framework. 1) We first collect multiple graph tasks, and unify them into a code-like format, along with task-specific textual data to form a graph instruction corpus. 2) Then, we perform graph instruction tuning to improve the ability of an LLM to solve graph reasoning and generation tasks. 3) Finally, we investigate multiple graph hallucination scenarios and optimize the LLM by preference alignment.}
 \label{fig:instructgraph_framework}
\end{figure*}

The skeleton is shown in Figure~\ref{fig:instructgraph_framework}, which can be decomposed into three modules, i.e., graph input engineering, graph instruction tuning, and graph preference aligning.

\subsection{Notation}

Suppose that there are $M$ graph tasks $\mathcal{D}=\{\mathcal{D}_1, \cdots\mathcal{D}_{M}\}$, and the corresponding dataset of each task can be denoted as $\mathcal{D}_j=\{(\mathcal{I}_i, \mathcal{G}_i, \mathcal{P}_i, \mathcal{A}_i)\}_{i=1}^{N_j}$, where $N_j$ denotes the number of examples of $\mathcal{D}_j$, $\mathcal{I}_i$ is the corresponding instruction~\footnote{We manually design the instruction for each dataset.}, $\mathcal{G}_i=(\mathcal{E}_i, \mathcal{R}_i, \mathcal{T}_i, \mathcal{S}_i)$ is the graph with one node (entity) set $\mathcal{E}_i$, one optional relation set $\mathcal{R}_i$, one edge (triple) set $\mathcal{T}_i$, and one optional textual property set $\mathcal{S}_i$, $\mathcal{P}_i$ is the optional passage 
, and $\mathcal{A}_i$ is the final answer~\footnote{Especially, the answer $\mathcal{A}_i$ can be not only an independent text but also one of $\mathcal{G}_i$ and $\mathcal{P}_i$, depending on the task paradigm.}.

\subsection{Graph Input Engineering}

The first challenge is how to align the graph to the text to meet the sequence interface of LLMs,
previous works solved this issue by using graph description~\cite{Ye2023Natural} or embedding fusion~\cite{Chai2023GraphLLM}, which may make the generated responses difficult to parse into actual graphs.

Inspired by current LLMs that can simultaneously understand and generate code, we introduce a \emph{structured format verbalizing} strategy to transform the graph into a simple code-like format.
Formally, given one task graph $\mathcal{G}_i\in\mathcal{D}_j$, 
we denote $M(\cdot)$ as the structured format verbalizer, and the original graph can be mapped into a sequence as $\mathcal{C}_i=M(\mathcal{G}_i)$.
For the fundamental format, all nodes (or entities) are listed as a sequence with variable \texttt{node\_list} (or \texttt{entity\_list}), while all edges (or triples) are listed as a sequence with variable \texttt{edge\_list} (or \texttt{triple\_list}).
For graphs that contain side information,
we can simulate the object-oriented language to express the node (or entity).
Take the graph in Figure~\ref{fig:instructgraph_all_task} as an example, the review text ``The film is nice.'' of the node ``User1'' can be expressed by ``User1.review=The film is nice.'', where ``.review'' can be replaced as the property name in the graph. 
Therefore, we can unify all graphs into a unified format to align with textual data.

\subsection{Graph Instruction Tuning}

As shown in Figure~\ref{fig:instructgraph_all_task}, we first define four different groups of graph-centric instruction tasks to bolster the ability of LLMs on the graph, including graph structure modeling, graph language modeling, graph generation modeling, and graph thought modeling. The first two groups are focused on graph reasoning, the third group is typical graph generation, and the last group aims at both graph reasoning and generation~\footnote{We only choose the first three groups of tasks for instruction tuning. The tasks from graph thought modeling are only used for the evaluation.}.
After graph input engineering, we can directly reuse the standard causal language modeling (CLM) objective to continually tune the LLM on such groups.
Formally, given one task dataset $\mathcal{D}_j=\{(\mathcal{I}_i, \mathcal{G}_i, \mathcal{P}_i, \mathcal{A}_i)\}_{i=1}^{N_j}$,
the LLM can be optimized by \emph{maximum likelihood} with:
\begin{equation}
\mathcal{L}(\mathcal{D}_j)=-\sum_{i=1}^{N_j}\log \pi_{\theta}(\mathcal{Y}_i=\mathcal{A}_i|\mathcal{X}_i),
\label{eqn:instruct_objective}
\end{equation}
where $\pi_{\theta}$ denotes the LLM with trainable parameters $\theta$, $\mathcal{Y}_i$ is the model output, $\mathcal{X}_i$ and $\mathcal{A}_i$ respectively represent the input sequence and reference label, which depends on the specific task definition.
Table~\ref{tab:graph_tasks} lists all groups of tasks and corresponding clusters to show the task definition, model input, and output.
Therefore, we can obtain an instruction-based graph LLM and named {\model}-INS.

\begin{table*}[t]
\centering
\resizebox{\linewidth}{!}{
\begin{tabular}{c|c|c|l|c}
\toprule
\bf Task Groups & \bf Task Clusters &\bf \makecell[c]{Task Definition} &\bf \makecell[c]{Task Input} & \bf Task Output \\
\midrule
\makecell[c]{Graph \\Structure \\Modeling} & \makecell[c]{Connection Detection, \\Cycle Detection, \\Hamilton Path, \\Bipartite Matching, \\Shortest Path, \\Degree Computing} & \makecell[l]{The tasks in this group aim to make LLMs better \\understand some basic graph structures. The \\input only contains nodes, directed or un-directed \\edges, and optional weights.} & \makecell[c]{$\mathcal{X}_i=[\mathcal{I}_i, \mathcal{C}_i]$} & \makecell[c]{$\mathcal{Y}_i=\mathcal{A}_i$} \\

\midrule
\multirow{6}*{\makecell[c]{\\\\\\\\\\Graph \\Language \\Modeling}} & \makecell[c]{Graph Caption \\Generation} & \makecell[l]{The task aims to generate a caption passage $\mathcal{P}_i$ \\to describe the graph $\mathcal{G}_i$.} & \makecell[c]{$\mathcal{X}_i=[\mathcal{I}_i, \mathcal{C}_i]$} & \makecell[c]{$\mathcal{Y}_i=\mathcal{P}_i$} \\
\cline{2-5}
& \makecell[c]{Graph Question \\Answering} & \makecell[l]{The task aims to reason on the whole graph $\mathcal{G}_i$ \\and find an entity as the final answer $\mathcal{A}_i\in\mathcal{E}_i$.} & \makecell[c]{$\mathcal{X}_i=[\mathcal{I}_i, \mathcal{C}_i, \mathcal{P}_i]$} & \makecell[c]{$\mathcal{Y}_i=\mathcal{A}_i$} \\
\cline{2-5}
& \makecell[c]{Graph Node \\Classification} & \makecell[l]{The task aims to classify the target node into pre-\\defined classes based on $\mathcal{G}_i$.} & \makecell[c]{$\mathcal{X}_i=[\mathcal{I}_i, \mathcal{C}_i, \mathcal{P}_i]$} & \makecell[c]{$\mathcal{Y}_i=\mathcal{A}_i$} \\
\cline{2-5}
& \makecell[c]{Graph Link \\Prediction} & \makecell[l]{The task aims to predict the relation between two \\given nodes based on $\mathcal{G}_i$.} & \makecell[c]{$\mathcal{X}_i=[\mathcal{I}_i, \mathcal{C}_i, \mathcal{P}_i]$} & \makecell[c]{$\mathcal{Y}_i=\mathcal{A}_i$} \\
\cline{2-5}
& \makecell[c]{Graph Relevance \\Inspection} & \makecell[l]{The task aims to detect whether the graph $\mathcal{G}_i$ is \\relevant to the passage $\mathcal{P}_i$, we have \\$\mathcal{A}_i\in\{\text{relevant}, \text{irrelevant}\}$.} & \makecell[c]{$\mathcal{X}_i=[\mathcal{I}_i, \mathcal{C}_i, \mathcal{P}_i]$} & \makecell[c]{$\mathcal{Y}_i=\mathcal{A}_i$} \\
\cline{2-5}
& \makecell[c]{Graph Collaboration \\Filtering} & \makecell[l]{The task aims to predict whether the target user \\prefers the target item based on the whole graph \\$\mathcal{G}_i$, the answer $\mathcal{A}_i$ can be set as a score.} & \makecell[c]{$\mathcal{X}_i=[\mathcal{I}_i, \mathcal{C}_i, \mathcal{P}_i]$} & \makecell[c]{$\mathcal{Y}_i=\mathcal{A}_i$} \\

\midrule
\multirow{2}*{\makecell[c]{Graph \\Generation \\Modeling}} & \makecell[c]{Knowledge Graph \\Generation} & \makecell[l]{The task aims to given a passage $\mathcal{P}_i$ that describes \\a piece of factual or commonsense information, \\the task aims to extract entities and relations from \\$\mathcal{P}_i$ to generate a graph $\mathcal{G}_i$.} & \makecell[c]{$\mathcal{X}_i=[\mathcal{I}_i, \mathcal{P}_i]$} & \makecell[c]{$\mathcal{Y}_i=\mathcal{C}_i$} \\
\cline{2-5}
& \makecell[c]{Structure Graph \\Generation} & \makecell[l]{The task aims to generate a graph to meet the \\structure information described in the passage $\mathcal{P}_i$.} & \makecell[c]{$\mathcal{X}_i=[\mathcal{I}_i, \mathcal{P}_i]$} & \makecell[c]{$\mathcal{Y}_i=\mathcal{C}_i$} \\
\midrule
\makecell[c]{Graph \\Thought \\Modeling} & \makecell[c]{Arithmetic \\Symbolic \\Robotic \\Logic} & \makecell[l]{The task aims to solve the general reasoning task \\in three think steps: 1) first find the question \\subject, 2) then generate a thought graph $\mathcal{G}_i$ to \\express the rationale and 3) finally output the result \\$\mathcal{A}_i$ based on the graph.} & \makecell[c]{$\mathcal{X}_i=\mathcal{I}_i$} & \makecell[c]{$\mathcal{Y}_i=[\mathcal{C}_i; \mathcal{A}_i]$} \\
\bottomrule
\end{tabular}
}
\caption{The overview of all groups of tasks.}
\vspace{-0.5em}
\label{tab:graph_tasks}
\end{table*}

\subsection{Graph Preference Alignment}


Recently, the NLP community has witnessed a significant decrease in hallucination through preference optimization~\cite{Ouyang2022Training, Zhao2023Beyond, Rafailov2023Direct, MacGlashan2023Interactive}. 
Following this,
we propose graph preference alignment to alleviate the hallucination of LLMs on the graph.
As depicted in Figure~\ref{fig:instructgraph_framework}, we intuitively design four typical hallucination circumstances for graph reasoning and generation and perform negative sampling for each graph task.

\noindent\paragraph{Hallucinations in Graph Reasoning}
Typically, the instruction-version LLM may be a strong instruction follower, 
yet, sometimes fall into hallucinations because of the erroneous input or lack of knowledge:
1) \emph{correct graph but wrong answer} means the LLM makes a wrong prediction even though the input is legal, 
2) \emph{unfactual graph but wrong answer} means the wrong answer caused by a graph with unfaithful semantics to external knowledge,
3) \emph{conflict graph but wrong answer} means there exists conflict information in the input graph, and 
4) \emph{missing graph but wrong answer} means that the input graph is missing some crucial information related to the answer.

To simulate the first circumstance, we can randomly choose a result from other examples to form a negative output $\mathcal{Y}_i^{-}$.
For the rest, we can randomly \emph{replace}, \emph{add}, or \emph{remove} some nodes (entities) or edges (triples) in the graph and construct a new input with the original instruction and passage. Therefore, the original answer can be viewed as the negative $\mathcal{Y}_i^{-}$ and the positive $\mathcal{Y}_i^{+}$ defined as ``Sorry, the input graph contains wrong information, so the question is unanswerable directly.''.

\noindent\paragraph{Hallucination in Graph Generation}
Graph generation is harder than reasoning because the LLM needs to output a complete and accurate code-like format sequence.
The following are three kinds of wrong-generated graphs: \emph{unfactual graph}, \emph{conflict graph} and \emph{missing graph}.
We can directly construct a wrong graph as the final output $\mathcal{Y}_i^{-}$ by performing \emph{replace}, \emph{add}, and \emph{remove} operators, which are similar to the graph reasoning.
The original graph is denoted as positive $\mathcal{Y}_i^{+}$.
Additionally, in cases where an incorrect answer is due to a faulty input, we may substitute the original input with an unrelated one from the dataset that doesn't affect the answer graph. The original answer graph is then considered as the negative output $\mathcal{Y}_i^{-}$.

We next use the DPO algorithm to reduce hallucination.
Specifically, given one instruction example $(\mathcal{X}_i, \mathcal{Y}_i^{+})$ and a corresponding negative $(\mathcal{X}_i, \mathcal{Y}_i^{-})$, we can define the preference model under the Bradley-Terry~\cite{Bradley1952Rank} 
as:
\begin{equation}
\begin{small}
\begin{aligned}
p_{\theta}(\mathcal{Y}_i^{+}>\mathcal{Y}_i^{-}|\mathcal{X}_i)=& \frac{1}{1+\exp\{r(\mathcal{Y}_i^{+}, \mathcal{Y}_i^{-}, \mathcal{X}_i)\}},\\
r(\mathcal{Y}_i^{+}, \mathcal{Y}_i^{-}, \mathcal{X}_i) =& - \beta\log\frac{\pi_{\theta}(\mathcal{Y}_i^{+}|\mathcal{X}_i)}{\pi_{\mathit{ref}}(\mathcal{Y}_i^{+}|\mathcal{X}_i)} \\
& + \beta\log\frac{\pi_{\theta}(\mathcal{Y}_i^{-}|\mathcal{X}_i)}{\pi_{\mathit{ref}}(\mathcal{Y}_i^{-}|\mathcal{X}_i)},
\label{eqn:preference_model_2}
\end{aligned}
\end{small}
\end{equation}
where $\beta$ is the balance factor, $p_{\theta}$ denotes the preference model, $\pi_{\theta}$ and $\pi_{\mathit{ref}}$ respectively denotes the policy and reference model, which can be initialized from instruction-version LLM.
Thus, we can optimize the LLM by \emph{maximum likelihood} with:
\begin{equation}
\begin{small}
\begin{aligned}
& \mathcal{J}({\pi_{\theta}, \pi_{\mathit{ref}}}) = -\mathbb{E}_{(\mathcal{X}_i, \mathcal{Y}_i^{+}, \mathcal{Y}_i^{-})\sim\mathcal{D}} \\
& \bigg[\log\sigma\big(
\beta\log\frac{\pi_{\theta}(\mathcal{Y}_i^{+}|\mathcal{X}_i)}{\pi_{\mathit{ref}}(\mathcal{Y}_i^{+}|\mathcal{X}_i)} - \beta\log\frac{\pi_{\theta}(\mathcal{Y}_i^{-}|\mathcal{X}_i)}{\pi_{\mathit{ref}}(\mathcal{Y}_i^{-}|\mathcal{X}_i)}
\big)
\bigg].
\end{aligned}
\end{small}
\label{eqn:preference_objective}
\end{equation}
We denote the policy $\pi_{\theta}$ as InstructGraph-PRE.

\begin{table*}[t]
\centering
\resizebox{\linewidth}{!}{
\begin{tabular}{c|cc | cc |cc | c}
\toprule
\textbf{Clusters} & \textbf{Tasks} & \textbf{Metrics} & \textbf{GPT-3.5} & \textbf{GPT-4} & \textbf{LLaMA2} &
\textbf{Vicuna} & \textbf{{\model}-INS} \\
\hline
\multirow{6}*{\textbf{Structure}} & Conn. Dect.  & ACC & 81.45 & 80.47 & 54.01 & 54.85 & \textbf{83.54} \\
& Cycle Dect.  & ACC & 59.02 & 61.44 & 50.79 & 52.88  & \textbf{91.10} \\
& Hami. Path & ACC & 21.03 & 29.10 & 1.23 & 1.23 & \textbf{34.80}\\
& Bipt. Match & ACC & 50.23 & 66.11 & 0.00 & 0.00 & \textbf{76.36} \\
& Shrt. Path & ACC & 38.99 & 49.03 & 0.00 & 0.00 & \textbf{66.29} \\
& Degree Comp. & ACC & 41.18 & \textbf{70.59} & 18.13 & 19.57 & 65.65 \\
\midrule
\multirow{5}*{\textbf{Caption}} & Wikipedia & BLEU & 91.99 & 93.85 & 77.15 & 82.94 & \textbf{95.81}  \\
& WebNLG & BLEU & \textbf{99.51} & 99.29 & 88.67 & 89.33 & 97.35 \\
& GenWiki & BLEU & 98.60 & \textbf{98.65} & 79.72 & 87.67 & 97.71 \\
& EventNA & BLEU & 62.66 & 61.75 & 53.39 & 75.52 & \textbf{81.64} \\
& Xalign & BLEU & 86.77 & 88.59 & 84.05 & 86.05 & \textbf{93.08}\\
\midrule
\multirow{4}*{\textbf{Graph QA}} & PathQSP & EM & 52.54 & 68.64 & 42.70 & 31.90  & \textbf{86.40} \\
& GrailQA & EM & 43.92 & 60.17 & 15.83 & 17.95 & \textbf{81.30}\\
& WebQSP & EM & 53.73 & 61.57 & 40.07 & 26.42 & \textbf{73.30}\\
& WikiTQ & EM & 49.02 & \textbf{60.78} & 29.94 & 35.76 & 47.82\\
\midrule
\multirow{5}*{\textbf{Node CLS}} & Cora & EM & 74.51 & 64.17 & 83.04 & 84.08 & \textbf{89.33} \\
& Citeseer & EM & 70.39 & \textbf{74.94} & 68.24 & 67.94 & 71.65\\
& Pubmed & EM & 74.63 & 77.16 & 79.78 & 80.18 & \textbf{81.09}\\
& Arxiv & EM & 70.59 & 74.51 & 45.50 & 57.75 & \textbf{81.50} \\
& Products & EM & 68.82 & 84.16 & 29.34 & 79.50  & \textbf{95.20} \\
\midrule
\multirow{3}*{\textbf{Link Pred.}} & Wikidata & Hits@1 & 43.73 & 62.94 & 10.75 & 10.38 & \textbf{96.52} \\
& FB15K-237 & Hits@1 & 60.34 & 66.88 & 0.00 & 0.00 & \textbf{98.91} \\
& ConceptNet & Hits@1 & 31.33 & 38.30 & 8.30 & 8.19 & \textbf{59.86} \\
\midrule
\textbf{Relevance} & Wikipedia & ACC & 94.40 & \textbf{100} & 69.27 & 68.12 & \textbf{100} \\
\midrule
\multirow{1}*{\textbf{RecSys}} & Amazon & Hits@1 & 27.09 & 59.77 & 44.40 & 16.40 & \textbf{78.80} \\
\midrule
\multirow{3}*{\textbf{IE}} & Wikipedia & F1 & 50.97 & 46.89 & 40.76 & 38.84 & \textbf{83.56} \\
& UIE & F1 & 24.41 & 26.22 & 20.21 & 26.11 & \textbf{76.82} \\
& InstructKGC & F1 & 21.44 & 21.86 & 19.26 & 16.6 & \textbf{38.98} \\
\midrule
\multirow{1}*{\textbf{Graph Gen.}} & NLGraph & F1 & 80.86 & 88.17 & 3.64 & 42.21 & \textbf{91.05} \\
\midrule
\multicolumn{3}{l|}{\textbf{Avg.}} & 59.45 & 66.76 & 41.65 & 46.06 & \textbf{79.84}\\
\bottomrule
\end{tabular}
}
\caption{Main results (\%) over multiple graph instruction tuning tasks under zero-shot settings. The number highlighted in bold denotes the best performance.}
\vspace{-0.56em}
\label{tab:all_instruct_results}
\end{table*}

\section{Experiments}

In this section, we perform extensive experiments to evaluate the effectiveness of {\model} over graph tasks and general NLP tasks.

\subsection{Implementation Settings}
We construct about 1.6M examples for graph instruction tuning and 100K examples for graph preference alignment.
In default, we choose LLaMA2-7B-HF
~\cite{Touvron2023Llama2} from HuggingFace\footnote{\url{https://huggingface.co/meta-llama}.} as the backbone.
The maximum length is set as $2048$.
The optimizer is AdamW. The learning rate is set to $5e-5$ with a decay rate of $0.1$ in the graph instruction tuning stage and will be changed to $5e-7$ in the graph preference alignment stage.
To accelerate the training\footnote{The implementation is referred to~\url{https://github.com/facebookresearch/llama-recipes}.}, we utilize FSDP~\cite{Zhao2023PyTorch} with CPU Offloading~\cite{Tsog2021Offloading}, FlashAttention~\cite{Dao2022FlashAttention}, and BFloat16 techniques, and utilize LoRA~\cite{Hu2022LoRA} to perform parameter-efficient learning with $rank=32$ and $lora\_\alpha=128$. 

\begin{figure*}[t]
\centering
\begin{tabular}{ccc}
\begin{minipage}[t]{0.33\linewidth}
    \includegraphics[width = 1\linewidth]{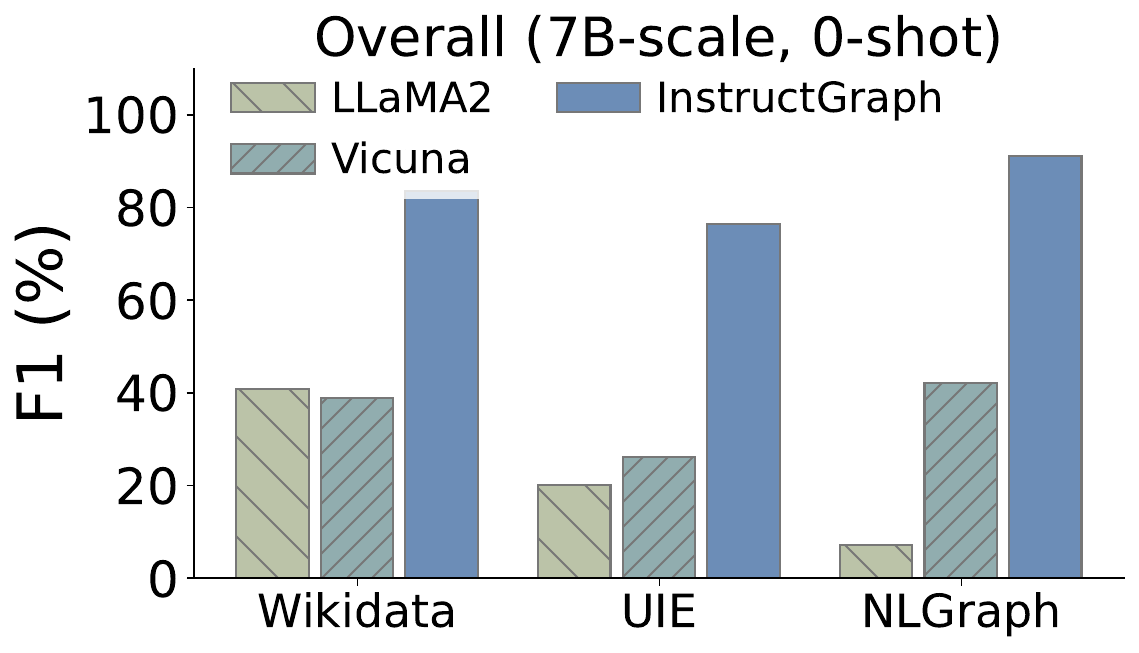}
\end{minipage}
\begin{minipage}[t]{0.33\linewidth}
    \includegraphics[width = 1\linewidth]{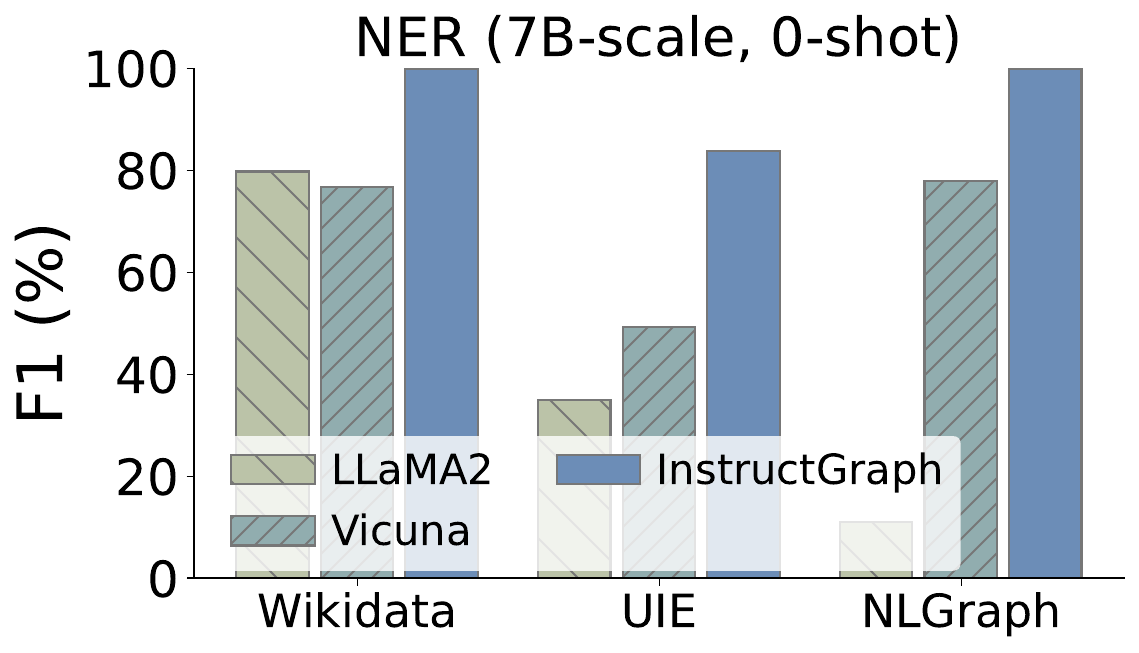}
\end{minipage}
\begin{minipage}[t]{0.33\linewidth}
    \includegraphics[width = 1\linewidth]{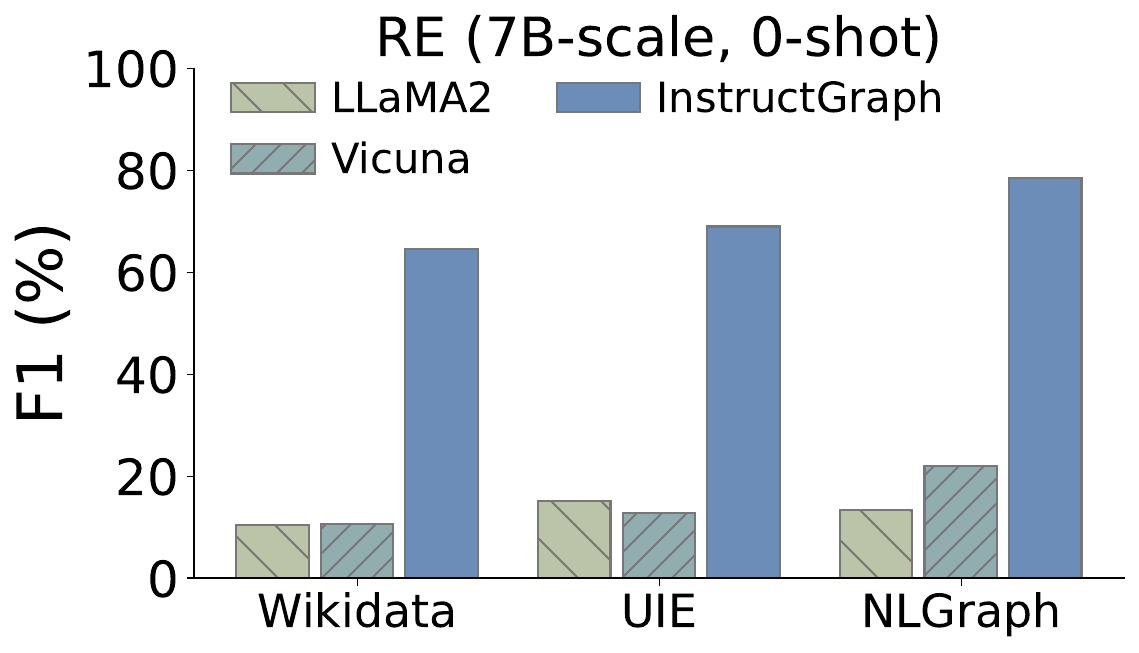}
\end{minipage}
\end{tabular}
\begin{tabular}{ccc}
\begin{minipage}[t]{0.33\linewidth}
    \includegraphics[width = 1\linewidth]{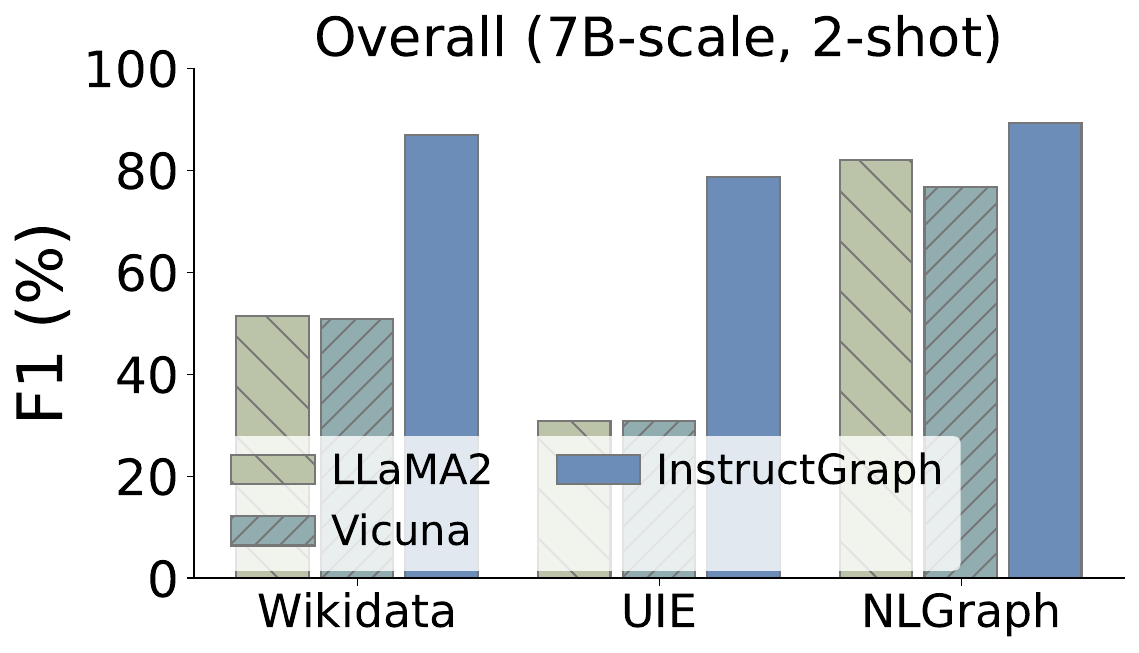}
\end{minipage}
\begin{minipage}[t]{0.33\linewidth}
    \includegraphics[width = 1\linewidth]{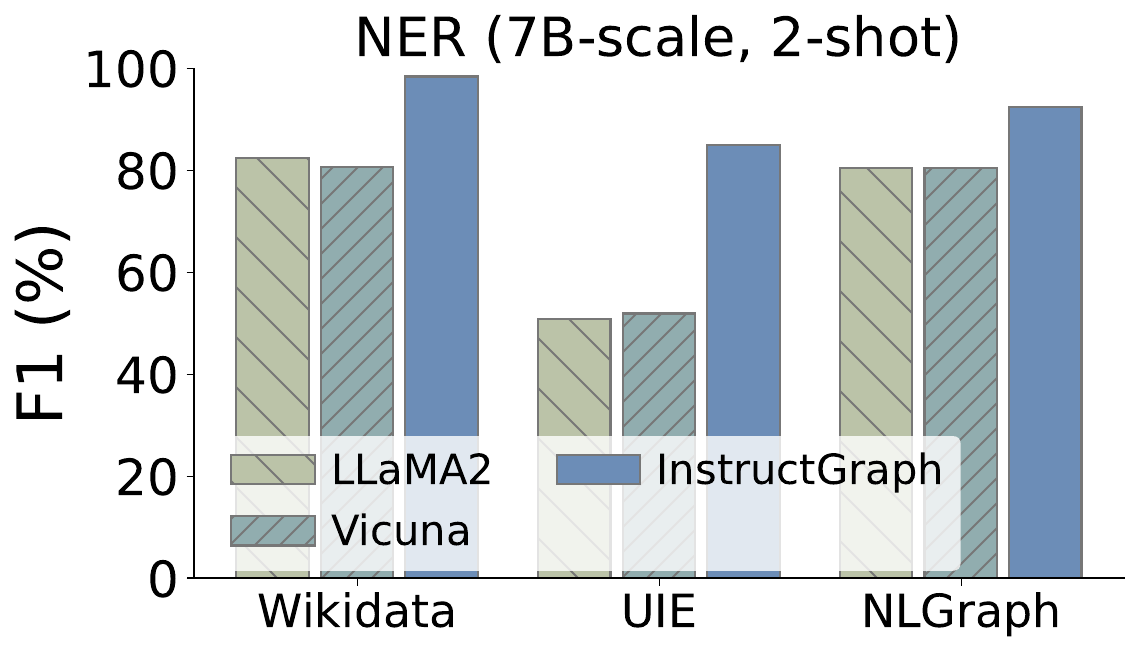}
\end{minipage}
\begin{minipage}[t]{0.33\linewidth}
    \includegraphics[width = 1\linewidth]{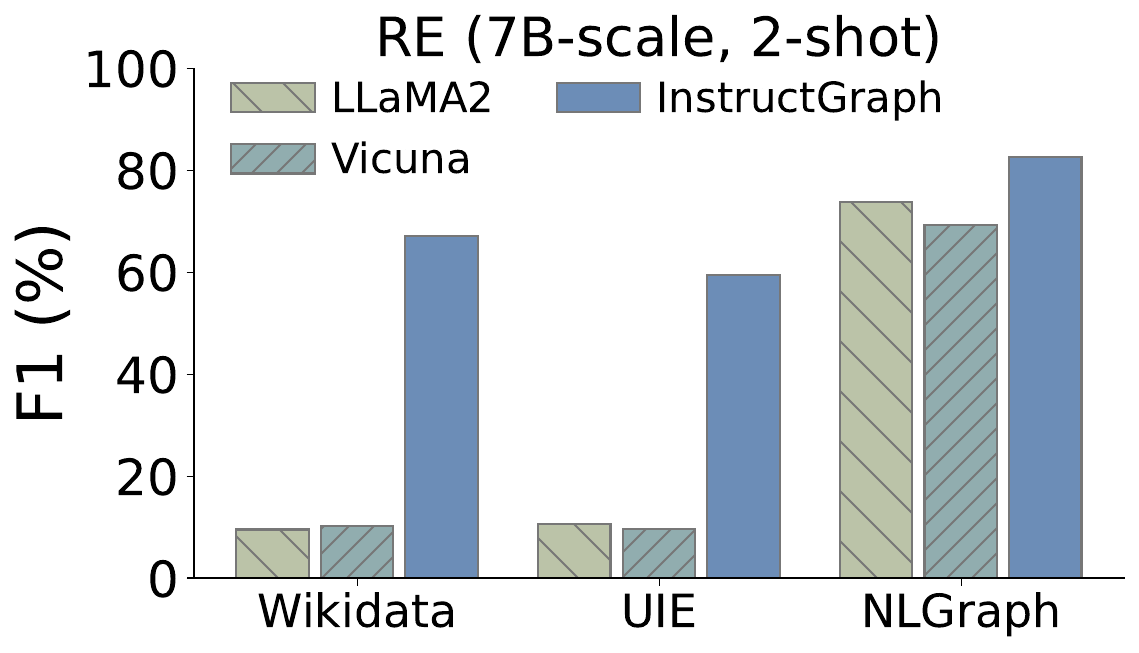}
\end{minipage}
\end{tabular}
\begin{tabular}{ccc}
\begin{minipage}[t]{0.33\linewidth}
    \includegraphics[width = 1\linewidth]{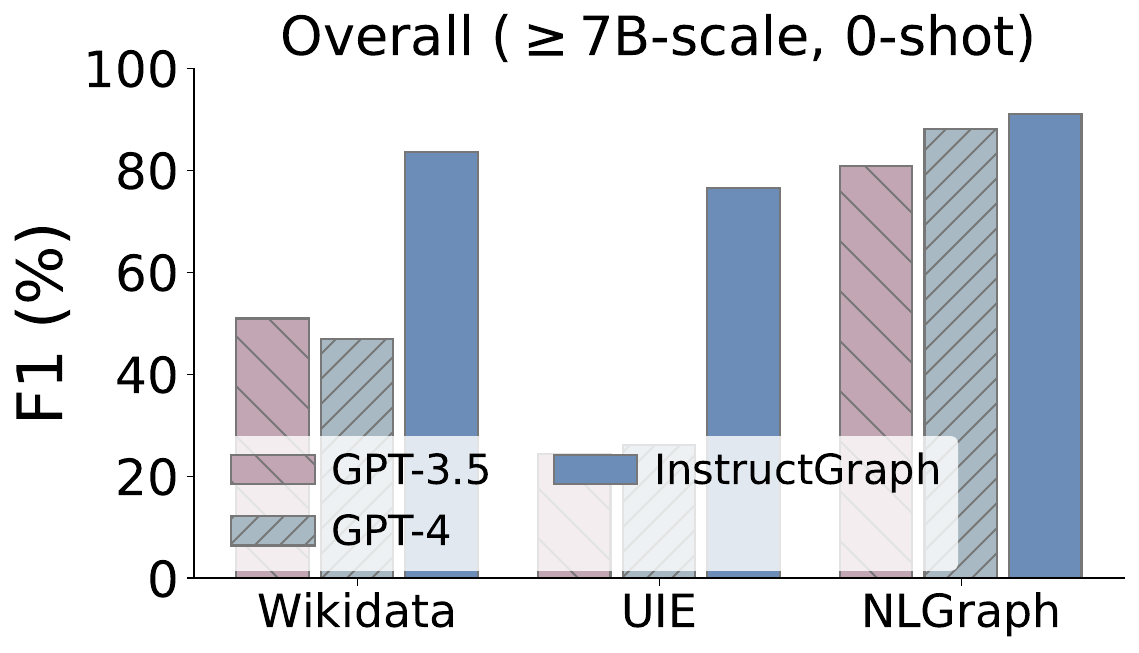}
\end{minipage}
\begin{minipage}[t]{0.33\linewidth}
    \includegraphics[width = 1\linewidth]{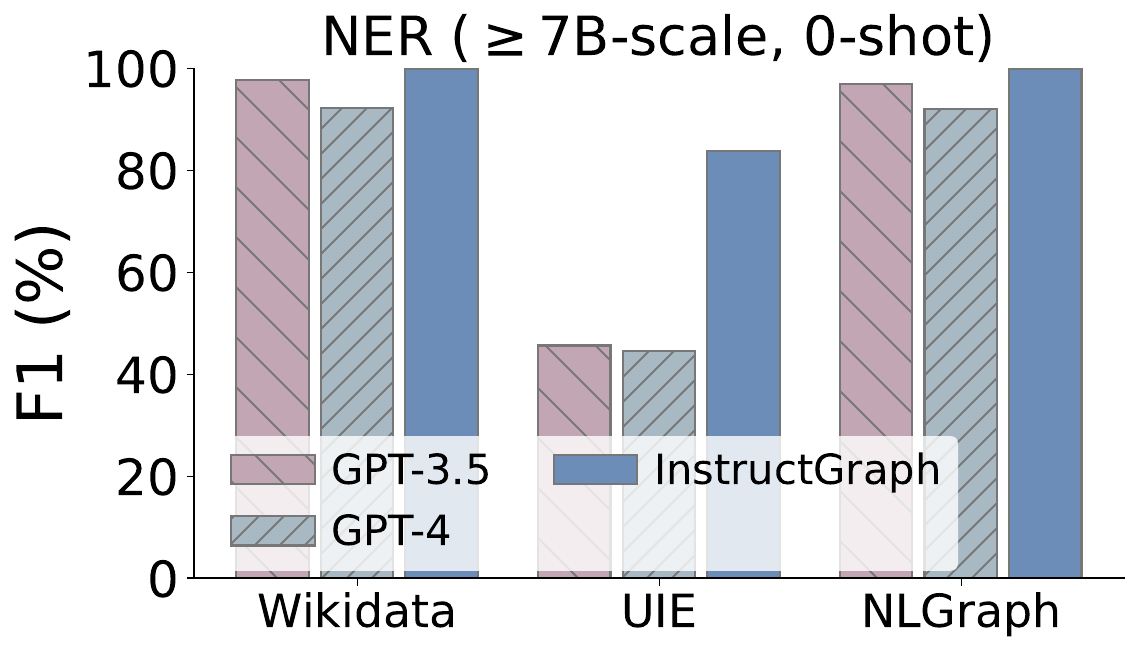}
\end{minipage}
\begin{minipage}[t]{0.33\linewidth}
    \includegraphics[width = 1\linewidth]{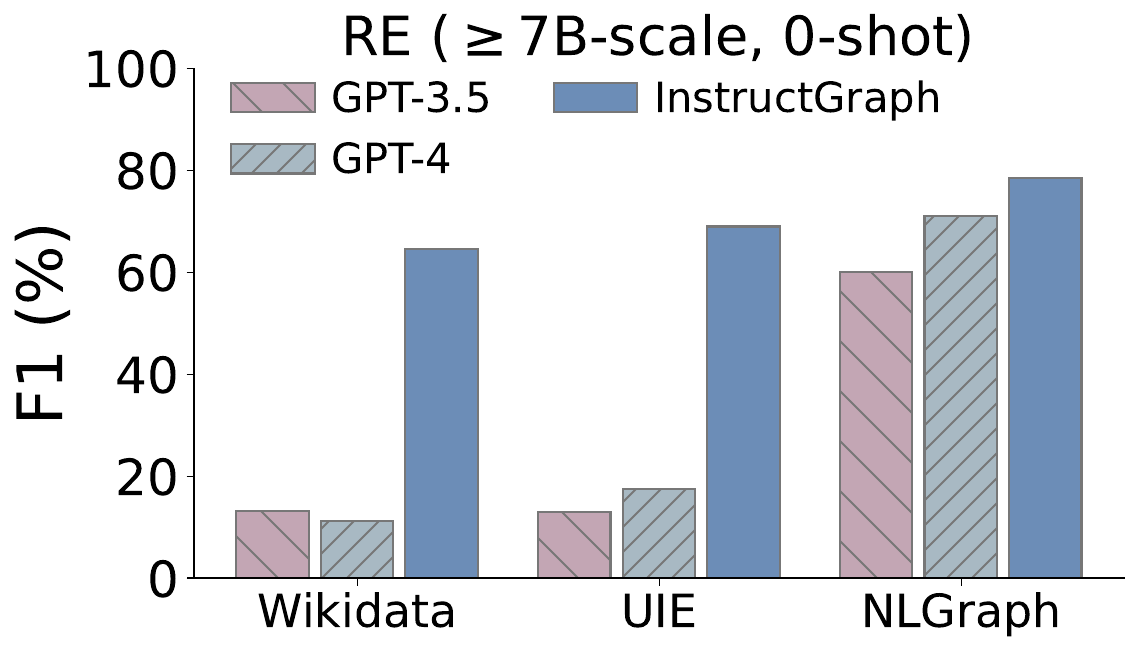}
\end{minipage}
\end{tabular}
\caption{Performance (\%) comparison with LLaMA2, Vicuna, GPT-3.5, and GPT-4 towards the overall graph, named entity recognition (NER), and relation extraction (RE) on graph generation tasks.}
\label{fig:all_generation_results}
\end{figure*}

\begin{table*}[t]
\centering
\begin{small}
\begin{tabular}{l|c|ccccc|c}
\toprule
\textbf{Methods (7B)} & \textbf{Is Align} &\textbf{Structure} & \textbf{Caption} &\textbf{Graph QA} & \textbf{Nodel CLS}  & \textbf{IE} & \textbf{Avg.} \\
\midrule
LLaMA2 & \XSolidBrush &38.64 &57.96 &70.70 &74.68 &37.40 &55.88 \\
Vicuna & \XSolidBrush &39.12& 62.37& 64.38& 77.63& 40.8& 56.86 \\
\textbf{\model-INS}  &\XSolidBrush &50.32& 81.15& 77.85& 83.16& 69.14& 72.32 \\
\textbf{\model-PRE}  &\Checkmark &\textbf{57.80} &\textbf{87.44}& \textbf{84.44}& \textbf{88.98}& \textbf{91.44}& \textbf{82.02} \\
\bottomrule
\end{tabular}
\end{small}
\caption{Main results (\%) over multiple graph preference tasks under zero-shot settings.}
\label{tab:graph_preference_results}
\end{table*}

\begin{table*}[t]
\centering
\resizebox{\linewidth}{!}{
\begin{tabular}{l|c|cccc|cc|c|c}
\toprule
\multirow{2}*{\textbf{Methods}} & \multirow{2}*{\textbf{Is Align}} & \multicolumn{4}{c|}{\textbf{HaluEval}} & \multicolumn{2}{c|}{\textbf{Anthropic-HH}} & \multirow{2}*{\textbf{TruthfulQA}} & \multirow{2}*{\textbf{Avg.}} \\
& & \textbf{Dialogue} & \textbf{General} &\textbf{QA} & \textbf{Abstract}  & \textbf{Harmless} & \textbf{Helpful} & &  \\
\midrule
GPT-3.5 & \Checkmark &\bf 72.40&\bf 79.44 &\bf 62.59 &\bf 58.53 & - & - &  47.50  & - \\
GPT-4 & \Checkmark & - & - & - & - & - & - & \bf 59.80 & - \\
\midrule
LLaMA2-7B & \XSolidBrush &43.99& 20.46& 49.60& 49.55 &54.28 &\bf 60.49  & 33.29  & 44.52 \\
Vicuna-7B & \XSolidBrush &46.35& 19.48& \textbf{60.34}& 45.62 &55.70 &58.71  & 30.10 & 45.19 \\
\textbf{\model-INS}  &\XSolidBrush &44.88& 21.35& 52.90& 51.10 &56.33 &59.10  & 35.35 & 45.86 \\
\textbf{\model-PRE}  &\Checkmark &\bf 47.03&\bf  21.61& 52.88&\bf  51.39&\bf 58.40 & 60.12  &\bf  35.77 &\bf 46.74  \\
\bottomrule
\end{tabular}}
\caption{Main results (\%) over multiple universal NLP preference tasks under zero-shot settings.
}
\label{tab:nlp_preference_results}
\end{table*}

\subsection{Main Results on Graph Instruction Tasks}

In this section, we exhaustively evaluate the {\model}-INS on multiple graph reasoning and generation tasks in zero-shot settings. 
We use a code-like format to unify all graphs and construct an instruction tuning test set.
Data statistics are shown in Table~\ref{tab:instructgraph_corpus}, and the details are shown in Appendix~\ref{app:instruction_corpus}.
To make a comparison with a similar scale LLM, we choose the widely-used LLaMA2-7B and Vicuna-7B as the open-source baseline.
In pursuit of investigating the performance level of InstructGraph in the era of AGI, we also choose GPT-3.5 (turbo)~\cite{Ouyang2022Training} and GPT-4~\cite{OpenAI2023GPT4} as strong baselines~\footnote{\url{https://platform.openai.com/}.}.

Table~\ref{tab:all_instruct_results} showcases the main results of graph reasoning and generation, we thus draw the following conclusions:
1) {\model}-INS achieves the best overall results $79.84\%$ and 
outperforms 
GPT-4 by $13.08\%$.
2) Compared with the same scale LLMs, our framework performs the best on all graph tasks, which shows that further instruction tuning over well-designed graph tasks can better improve the reasoning and generation ability.
3) For the tasks Degree Computing, WebNLG, GenWiki, WikiTQ, and Citseer, {\model}-INS underperforms GPT-3.5 and GPT-4. Since the LLMs with large-scale parameters have stored more similar knowledge. 
Despite this, {\model}-INS still exhibits approximately 10\% better performance on other reasoning tasks.

Additionally, we also expect to delve into whether {\model}-INS achieves the improvement on graph generation tasks, 
We choose two external manners to evaluate the results:
1) \emph{NER} denotes named entity recognition, and 2) \emph{RE} denotes relation extraction.
As shown in Figure~\ref{fig:all_generation_results}, we visualize the comparison performances on three graph generation tasks, where \emph{Wikidata} and \emph{UIE} belong to knowledge graph construction and \emph{NLGraph} focus on structure graph generation.
We observe that:
1) {\model}-INS can bring significant improvement for LLaMA2 and Vicuna, indicating the graph generation ability encompasses NER and RE.
2) We also integrate all baselines with the 2-shot exemplars,
the results illustrate that the performance of {\model}-INS is consistently the highest.
3) RE is more challenging to NER 
because it involves understanding the semantics of generated nodes (entities) and making decisions on their relation or weight.
Despite this, the improvement of RE is larger than NER, which signifies that graph-specific optimization can better empower the LLM in constructing triples.

\begin{table*}[t]
\centering
\resizebox{\linewidth}{!}{
\begin{tabular}{l|ccc|cc|cc|cc}
\toprule
\multirow{3}*{\textbf{Methods (7B)}} & \multicolumn{3}{c|}{\textbf{Arithmetic}} & \multicolumn{2}{c|}{\textbf{Symbolic}}  & \multicolumn{2}{c|}{\textbf{Robotic}} & \multicolumn{2}{c}{\textbf{Logic}} 
\\
& \bf GSM8K & \bf SVAMP & \bf AQuA & \bf Letter & \bf Coin & \bf Termes & \bf Floortile & \bf ProofWriter & \bf FOLIO \\
& (4-shot) & (4-shot) & (4-shot) & (4-shot) & (4-shot) & (4-shot) & (4-shot) & (4-shot) & (4-shot) \\
\midrule
LLaMA2 w/. CoT & 11.89 & 23.30 & 18.60 & 0.00 & 0.00 & 0.00 & 0.00 & 30.64 & 32.40 \\
Vicuna w/. CoT & 14.33 & 24.19 & 17.80 & 1.50 & 0.00 & 0.00 & 0.00 & 28.77 & 33.15 \\
\textbf{\model-INS} w/. CoT & \textbf{17.52}  & \textbf{28.80} & \textbf{22.33} & \textbf{8.70} & \textbf{6.20} & \textbf{30.00} & \textbf{50.00} & \textbf{55.80} & \textbf{41.68} \\
\midrule
LLaMA2 w/. GTM & 14.38 & 23.10 & 20.13 & 2.00 & 0.00 & 0.00 & 0.00 & 33.19 & 34.80 \\
Vicuna w/. GTM & 15.10 & 24.84 & 19.60 & 1.50 & 0.00 & 0.00 & 0.00 & 31.50 & 36.19 \\
\textbf{\model-INS} w/. GTM & \textbf{19.46} & \textbf{27.10} & \textbf{23.80} & \textbf{7.40} & \textbf{9.40} & \textbf{30.00} & \textbf{50.00} & \textbf{52.77} & \textbf{43.06} \\
\bottomrule
\end{tabular}}
\caption{Results (\%) on thought planning tasks in few-shot scenarios.}
\label{tab:planning_results}
\end{table*}

\subsection{Main Results on Graph Preference Tasks}

We next explore whether 
{\model} can reduce the graph hallucination problem. We sample a few tasks from the corresponding cluster to build a hallucination testing set, including structure, caption, graph question answering, and node classification. 
The data statistics are shown in Table~\ref{tab:instructgraph_corpus}, and the details are shown in Appendix~\ref{app:preference_corpus}.
Specifically, each example consists of a correct answer and a wrong answer, we calculate the LLM's perplexity (PPL) on these answers and choose the option with the lowest PPL score as the preference results.
Therefore, the accuracy metric can reflect the performance of hallucination mitigation.

As shown in Table~\ref{tab:graph_preference_results}, we choose LLaMA2, Vicuna, and two variants of {\model} to make a comparison. 
{\model-INS} outperforms LLaMA2 and Vicuna by $16.44$\% and $15.46$\%, respectively, demonstrating that our framework with only graph instruction tuning can solve the preference tasks better.
This indicates that injecting task-related knowledge into the LLM's intrinsic parameter can be one of the significant factors for hallucination reduction.
Furthermore, {\model-PRE} significantly enhances the instruction version model by about 10\%, demonstrating that well-designed preference optimization can hit the upper boundary and endow the LLM with the ability to alleviate the pitfalls of hallucination.

We also delve into whether the preference optimization on the graph data hinders the effectiveness in the general domains.
To reach this goal, we choose three external preference and hallucination tasks. 1) HaluEval~\cite{Li2023HaluEval}~\footnote{\url{https://github.com/RUCAIBox/HaluEval}.} focuses on hallucination evaluation in dialogue, general understanding, question answering, and text summarization (abstract). 2) TruthfulQA~\cite{Lin2022TruthfulQA}~\footnote{\url{https://github.com/sylinrl/TruthfulQA}.} aims to test the factuality of LLMs on knowledge-intensive tasks. We choose MC1 as the test.
3) Anthropic-HH~\cite{Bai2022Training}~\footnote{\url{https://github.com/anthropics/hh-rlhf}.} has released the evaluation set for both harmless and helpful perspective.
For these tasks, we do not perform task-specific fine-tuning to show the zero-shot performance. Results in Table~\ref{tab:nlp_preference_results} showcase that our framework occasionally outperforms the sample scale baselines on some tasks, 
which meets our desiderata.

\subsection{Effectiveness of Thought Planning}

Recall the graph instruction tuning, 
we are eager for the LLM to solve
the thought planning tasks, including arithmetic, symbolic, robotic, and logic.
We design two few-shot scenarios:
1) \emph{Chain-of-Thought (CoT)} directly sampling few-shot exemplars with manually annotated sequence rationales to form a prompt.
2) \emph{Graph Thought Modeling (GTM)} decomposes the sequence rationale into three stages, i.e., finding topic entities or keywords, building a graph to express the thought, and outputting the final answer.
The comparison results are depicted in Table~\ref{tab:planning_results}, and we can observe that {\model}-INS achieves the best performance when elicited by CoT and GTM prompts.
In addition, GTM sometimes performs below expectations in the tasks of SVAMP, Letter, and ProofWriter. We believe that these tasks are difficult to express using an explicit graph to convey the thinking process.

\subsection{Performance on General NLP Tasks}

We next evaluate the performance of {\model} on the general NLP tasks.
We choose Big-Bench-Hard (BBH)~\cite{Suzgun2023Challenging} and Massive Multitask Language Understanding (MMLU)~\cite{Hendrycks2021Measuring} benchmarks with few-shot exemplars to perform reasoning.
As shown in Table~\ref{tab:universal_nlp_results}, 
even though these tasks do not belong to graph domains, we can still obtain competitive results compared with other same-scale open-source LLMs.

\begin{table}[t]
\centering
\begin{small}
\begin{tabular}{l|cc}
\toprule
\multirow{2}*{\textbf{Methods}} & \textbf{BBH} & \textbf{MMLU} \\
& (3-shot) & (5-shot) \\
\midrule
GPT-3.5 & - & 70.00  \\
GPT-4 & - & 86.40  \\
\midrule
MPT-7B & 31.00 & 26.80 \\
Falcon-7B & 28.00 & 26.20 \\
LLaMA-7B & 30.30 & 35.10 \\
LLaMA2-7B & 32.58 & 45.65 \\
Vicuna-7B & 31.54 & 50.34 \\
\textbf{\model-INS} & \textbf{33.06} & \textbf{51.62}  \\
\bottomrule
\end{tabular}
\end{small}
\caption{Results (\%) over multiple general NLP tasks under few-shot in-context learning settings.
}
\label{tab:universal_nlp_results}
\end{table}

\section{Analysis}


\subsection{Parameter-Efficient Learning Study}

\begin{figure}[t]
\centering
\includegraphics[width=\linewidth]{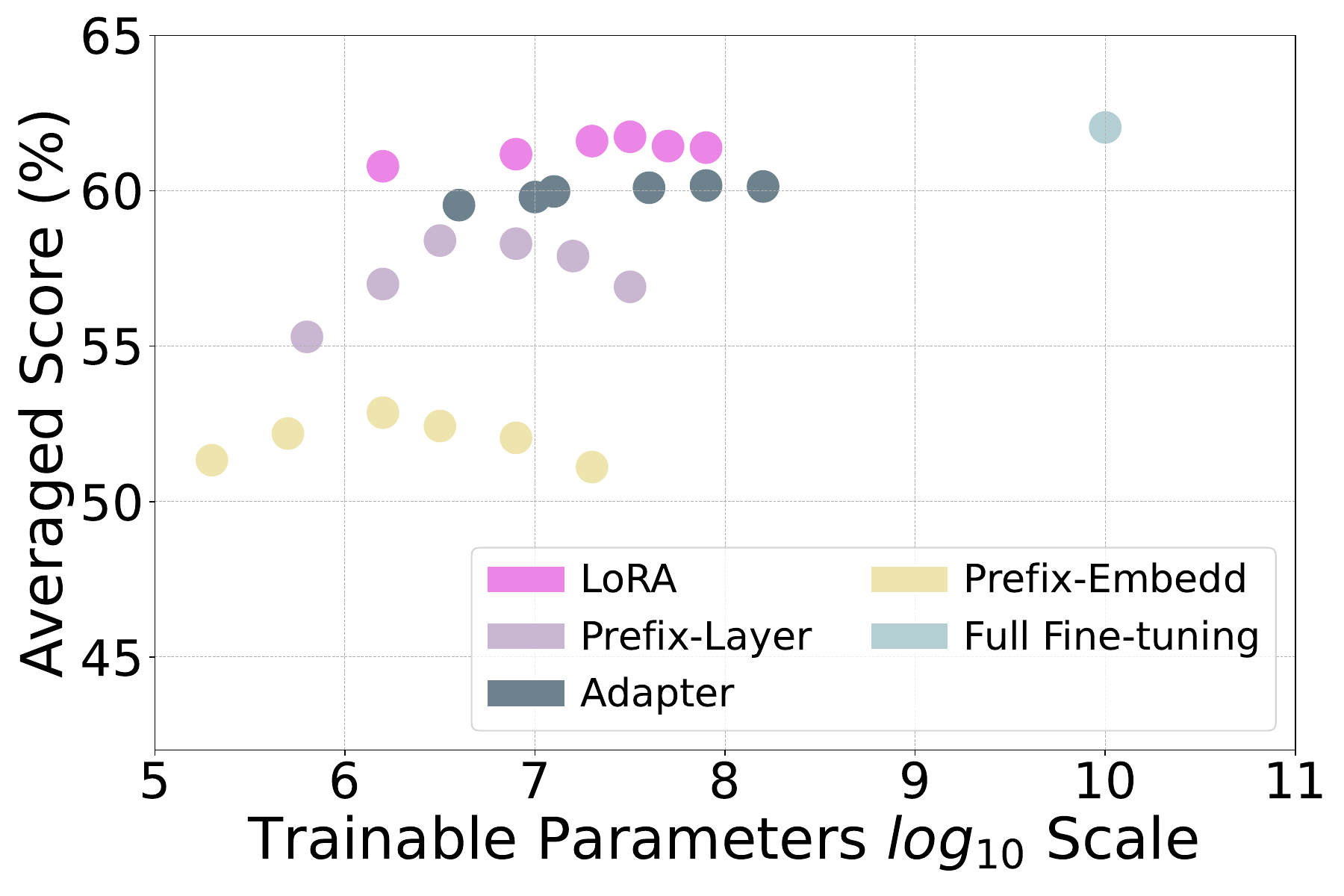}
\caption{Results (\%) of balance between trainable parameters and performances over graph tasks.}
 \label{fig:balance}
\end{figure}

To accelerate the training speed and reduce memory usage under the limitation of sources, we leverage parameter-efficient learning (PEL) techniques to equip the original LLM with only a few trainable parameters.
To study the choice of different PEL methods, we compare LoRA with other PEL methods, such as Prefix-tuning~\cite{Li2021Prefix}~\footnote{Prefix-Embedd: only tune the input embeddings layer; Prefix-Layer: tune each transformer layer.}, and Adapter~\cite{Houlsby2019Parameter}.
For each method, we choose six different scales and perform graph instruction tuning over 10\% training data.
The balance between trainable parameters and averaged results is visualized in Figure~\ref{fig:balance}. We can see that LoRA can achieve the best performance and is similar to full fine-tuning regardless of the scale of trainable parameters.

\begin{table}[t]
\centering
\begin{small}
\begin{tabular}{l|cccc}
\toprule
\textbf{Methods} & \textbf{PathQSP} & \textbf{WebNLG} & \textbf{CoRA} & \textbf{UIE} \\
\midrule
\multicolumn{5}{l}{\emph{GPT-4}} \\
\midrule
Template & 58.20 & 96.13 & 58.58 & 0.00 \\
Code Format  & 68.64 &  99.29 &  64.17 & 26.22 \\
\midrule
\multicolumn{5}{l}{\emph{LLaMA2}} \\
\midrule
Template & 20.36 & 59.15  & 27.44  & 0.00 \\
Code Format  & 42.70 & 88.67  &  83.04 & 20.21 \\
\bottomrule
\end{tabular}
\end{small}
\caption{Results (\%) comparison with different prompt engineering during the inference.
}
\label{tab:code_format_effectivness}
\end{table}

\begin{table}[t]
\centering
\begin{small}
\resizebox{\linewidth}{!}{
\begin{tabular}{l|ccc}
\toprule
\textbf{Baselines} & \textbf{Graph QA} & \textbf{Node CLS} & \textbf{IE} \\
\midrule
\multicolumn{4}{l}{\textit{Graph Instruction Testing}} \\
\midrule
\textbf{\model-INS} & \textbf{72.21} & \textbf{83.75} & \textbf{66.45} \\
w/. only GSM & 71.89 & 83.04 & 63.77 \\
w/. only GLM & 69.32 & 78.40 & 66.13 \\
w/. only GGM & 72.09 & 83.66 & 39.10 \\
w/. only GTM & 69.30 & 81.90 & 66.33 \\
\midrule
\multicolumn{4}{l}{\textit{Graph Preference Testing}} \\
\midrule
\textbf{\model-PRE} & \textbf{84.44} & \textbf{88.98} & \textbf{91.44} \\
w/o. only unfactual & 82.10 & 84.52 & 84.33 \\
w/o. only conflict & 83.70 & 85.17 & 81.11 \\
w/o. only missing & 79.35 & 83.55 & 78.40 \\
w/o. ALL & 77.85 & 83.16 & 69.14 \\
\bottomrule
\end{tabular}
}
\end{small}
\caption{Average performance (\%) of all tasks in each cluster when comparing different ablation versions.
GSM, GLM, GGM, and GTM denote graph structure modeling, graph language modeling, graph generation modeling, and graph thought modeling, respectively.
w/o. ALL equals to {\model}-INS.
}
\label{tab:ablation_study}
\end{table}

\subsection{Effectiveness of Code Format Graph}

In this part, we evaluate the use of the structured format verbalizer when aligning the graph structure to the textual LLM.
We choose four classic graph reasoning and generation tasks, i.e., PathQSP, WebNLG, CoRA, and UIE.
To compare with the structured format verbalizer, we directly choose the heuristic template introduced by InstructGLM~\cite{Ye2023Natural} to describe each path in the graph. For example, the path ``($e_1$, $r_1$, $e_2$), ($e_2$, $r_2$, $e_3$)'' can be formulated as ``$e_1$ is connected with $e_3$ within tow hops through $e_2$, and featured relations $r_1$ and $r_2$''.
We use this template to prompt GPT-4 and LLaMA2 to show the performance.
The results in Table~\ref{tab:code_format_effectivness} demonstrate that our structured format verbalizer outperforms traditional templates in all tasks. Especially, the LLM with traditional templates cannot support graph generation, while the structured format verbalizer can reach this goal.

\begin{figure*}[t]
\centering
\begin{tabular}{ccc}
\begin{minipage}[t]{0.49\linewidth}
    \includegraphics[width = \linewidth]{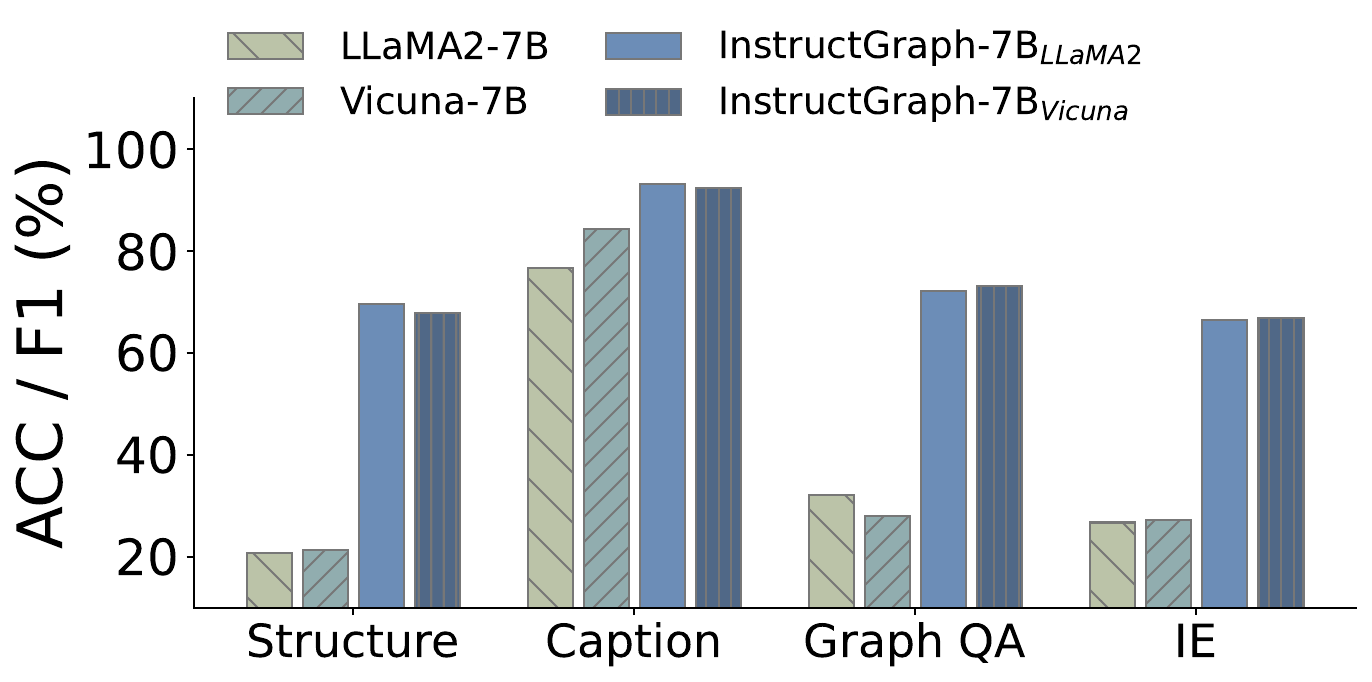}
\end{minipage}
\begin{minipage}[t]{0.49\linewidth}
    \includegraphics[width = \linewidth]{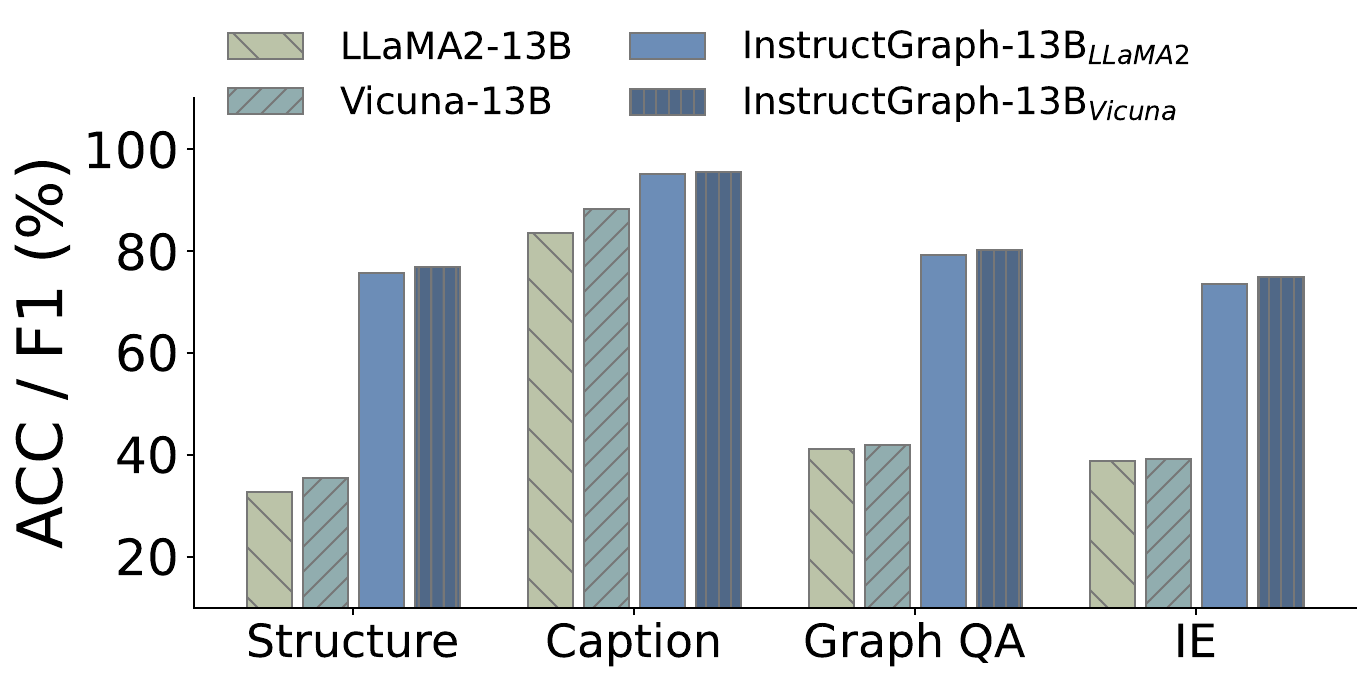}
\end{minipage}
\end{tabular}
\caption{Effectiveness (\%) of {\model} towards different scales and backbones.}
\label{fig:different_backbones}
\end{figure*}

\begin{table*}
    \centering
    \small
    \vspace{5pt}
    \resizebox{\linewidth}{!}{
    \begin{tabular}{p{8cm} | p{3.5cm} | p{3.5cm} | p{3.5cm} }
    \toprule
        \textbf{Instruction (Graph Caption)} & \textbf{LLaMA2 Output} & \textbf{{\model} Output} & \textbf{Reference}  \\
        \midrule
        You are a good graph reasoner. Give you a graph language that describes a graph structure and node information. You need to understand the graph and the task definition and answer the question.
        $\backslash$n```$\backslash$nGraph[name='wikipedia-knowledge-graph'] {$\backslash$n    entity\_list = ['clown beetle', 'euspilotus cribrum', 'america', 'the family', 'species', 'histeridae'];$\backslash$n    triple\_list = [('euspilotus cribrum' -> 'species')[relation='taxon rank'], ('the family' -> 'america')[relation='country']];$\backslash$n}$\backslash$n```$\backslash$n Task definition: given a knowledge graph with all entities and structure triples representing factual and commonsense knowledge. Please leverage this graph to generate an encyclopedia passage. Note that do not list all knowledge in a running account.$\backslash$n Q: Please generate an encyclopedia passage for the knowledge graph.$\backslash$n A:

        & The relation between Euspilotus cribrum and species is 'taxon rank', and the family country is America.
        \textcolor{red}{[\textbf{Human Evaluation}]: \emph{This answer directly states each fact triple from the given graph separately. It is accurate but not natural and satisfied as a whole.}}
        & Euspilotus cribrum is a species of clown beetle in the family Histeridae which is found in North America.
        \textcolor{blue}{[\textbf{Human Evaluation}]: \emph{This answer naturally describes the factual knowledge. Overall, it is very similar to an encyclopedia document and makes the reading experience better.}}
        & Euspilotus cribrum is a species of clown beetle in the family Histeridae. It is found in North America.  \\

        \midrule
        \textbf{Instruction (Graph Generation)} & \textbf{LLaMA2 Output} & \textbf{{\model} Output} & \textbf{Reference}  \\
        \midrule

        You are a good graph generator. You need to understand the task definition and generate a graph language to answer the question. 
        $\backslash$n Task definition: given a passage, extract some corresponding structure factual triples from the passage to form a knowledge graph, and generate a graph language to describe the graph. Note that: 1) the graph is a directed graph and the name is 'knowledge-graph'. 2) The entities must be mentioned in the passage. 3) The candidate relation type must be one of 'publication date', 'country of origin', 'publisher', 'performer'. 4) The generated graph language should be a code-like structure, and the skeleton format can be expressed as the following:$\backslash$n
        ```$\backslash$n Graph[name='knowledge-graph'] \{$\backslash$n entity\_list = ['xxx', $\cdots$];$\backslash$n    triple\_list = [('xxx' -> 'xxx')[relation='xxx'], $\cdots$];$\backslash$n \}$\backslash$n```$\backslash$n Passage: "'Bluesman' is the 12th original album of Tak Matsumoto, the guitarist of the Japanese music group B'z. It was released by VERMILLION RECORDS on September 2, 2020.".$\backslash$n Q: Given you a passage, please generate a corresponding knowledge graph.$\backslash$n A:
        & 
        Graph[name='knowledge-graph'] \{$\backslash$n entity\_list = ['Bluesman', 'Tak Matsumoto', 'B'z'];$\backslash$n    triple\_list = [('Bluesman' -> 'Tak Matsumoto) [relation='performer'], $\backslash$n ('Bluesman' -> 'B'z') [relation='performer'], $\backslash$n (Tak Matsumoto' -> 'B'z\') [relation='performer'],$\backslash$n ('Tak Matsumoto' -> 'VERMILLION RECORDS') [relation='publisher']];\}
        \textcolor{red}{[\textbf{Human Evaluation}]: \emph{This answer sometimes contains wrong triples or missing some entities.}}
        & 
        Graph[name=\'knowledge-graph\'] \{$\backslash$n    entity\_list = [\'Bluesman\', \'Tak Matsumoto\', \'B'z\', \'VERMILLION RECORDS\', \'September 2, 2020\'];$\backslash$n    triple\_list = [(\'Bluesman\' -> \'September 2, 2020\') [relation=\'publication date\'], (\'Bluesman\' -> \'Tak Matsumoto\') [relation=\'performer\'], (\'Bluesman\' -> \'VERMILLION RECORDS\') [relation=\'publisher\'], (\'Bluesman\' -> \'Japan\') [relation=\'country of origin\']];$\backslash$n\}
        \textcolor{blue}{[\textbf{Human Evaluation}]: \emph{This answer looks satisfying.}}
        & 
        Graph[name=\'knowledge-graph\'] \{$\backslash$n    entity\_list = [\'country of origin\', \'performer\', \'publication date\', \'Bluesman\', \'publisher\'];$\backslash$n    triple\_list = [(\'Bluesman\' -> \'Japan\') [relation=\'country of origin\'], (\'Bluesman\' -> \'VERMILLION RECORDS\') [relation=\'publisher\'], (\'Bluesman\' -> \'September 2, 2020\') [relation=\'publication date\'], (\'Bluesman\' -> \'Tak Matsumoto\') [relation=\'performer\']];$\backslash$n\}
        \\
        \bottomrule
    \end{tabular}
    }
    \caption{Human evaluation for the generation of LLaMA2 and {\model}.}
    \label{tab:case_study}
\end{table*}

\subsection{Ablation Study}

In this section, we focus on the ablation study to show how much each 
component contributes to performance.
We choose three clusters for the test, i.e., Graph QA, Node CLS, and IE.
For the graph instruction testing, we validate the effectiveness of each modeling task, and the test set is from the instruction corpus. For the graph preference testing, we evaluate three hallucination sampling strategies, including \emph{unfactual graph}, \emph{conflict graph}, and \emph{missing graph}, the test set is from the preference corpus.

As shown in Table~\ref{tab:ablation_study}, the results illustrate that the performance drops when removing one of these components.
For the instruction tuning testing, we can observe that graph language modeling plays a significant role in Graph QA and Node CLS clusters, while graph generation modeling is beneficial to the performance of IE.
For the preference testing, we can see that the performance of w/o. \emph{missing graph} drops significantly, indicating that the major factor of hallucination is the lack of key information in the input graph or generated graph.

\subsection{Effectiveness of Different Backbones}

To investigate whether the proposed {\model} can consistently improve the graph reasoning and generation ability with different LLMs, we select LLaMA2-7B, LLaMA2-13B, Vicuna-7B, and Vicuna-13B as the start checkpoints.
To make the experiment efficient, we randomly choose 10\% training data to perform graph instruction tuning and make a comparison with the corresponding vanilla LLMs.
Results in Figure~\ref{fig:different_backbones} show that {\model} can consistently achieve substantial improvement for arbitrary backbones and scales.
Additionally, we observe that Vicuna has better performance than LLaMA2 initially. However, after graph instruction tuning, this trend is reversed. Upon further analysis, we find that both LLaMA2 and Vicuna were re-optimized based on LLaMA~\cite{Touvron2023LLaMA}. 
Vicuna's optimization involves using supervised fine-tuning (SFT) to inject domain knowledge with massive conversation data into LLaMA. Meanwhile, LLaMA2 focuses on refactoring the model architecture and pre-training strategy to improve the model's versatility. 
Thus, Vicuna may have a better ability to understand instructions than LLaMA2.
Despite this, LLaMA2 can be the better starting checkpoint for boosting LLMs on graph reasoning and generation tasks with parameter updates.

\subsection{Human Evaluation}

We end this section with a case study to demonstrate the performance of LLMs when solving graph reasoning and generation tasks.
We choose LLaMA2 (7B) to make a comparison and respectively choose one example from graph caption generation and knowledge graph generation. For the answer, we perform a human evaluation to estimate the effectiveness of {\model}.
As shown in Table~\ref{tab:case_study}, {\model} can outperform all the baselines.
Specifically, compared with LLaMA2, {\model} can generate more natural and readable captions to describe factual information. For the graph generation, {\model} can provide accurate entities and triples.

\section{Related Work}


\subsection{LLMs for Graph Learning}

A series of works have studied how to leverage  LLMs to solve graph-centric tasks~\cite{Jin2023Large}, which can be decomposed into the following categories:
1) Prompt engineering. A series of works aims to design the interface to elicit the LLM to better understand and reason on the graph~\cite{Ye2023Natural, Han2023PiVe, Zhang2023LLM4DyG, Zhang2023Graph, Kim2023KGGPT, Wang2023Boosting, Luo2023Reasoning, Wang2023Can, Guo2023GPT4Graph, Zhao2023GraphText}.
2) Boosting LLMs with trainable GNNs. This kind of method focuses on enhancing the LLMs with trainable GNNs which can capture the arbitrary scale of the graph~\cite{Zhang2022GreaseLM, Chai2023GraphLLM, Tang2023GraphGPT, Zhao2023GIMLET, Tian2023Graph, Qin2023Disentangled}.
3) Instruction tuning over graph data. Similar to ours, \citet{Xu2023Symbol, Jiang2023StructGPT, Fang2023Mol, Zeng2023Interactive} directly collect some graph or symbol data to form an instruction corpus, and then continually pre-train the LLM. Different from them, our {\model} further empowers the LLM by graph instruction tuning with the code-like universal format and well-designed hallucination alleviation strategy by preference alignment.

\subsection{Hallucination in LLMs}
Recent works have studied that hallucination may degrade the performance of LLMs when performing instruction-follow inference.
LLMs usually generate seemingly plausible answers, which is called hallucination~\cite{Ji2023Survey, Zhang2023Siren}.
The phenomenon of hallucination encompasses fabricating erroneous user input, unfaithful for previously generated context, and unfactual for external knowledge and commonsense.
To estimate  hallucination, 
\citet{Kryscinski2020Evaluating, Li2023HaluEval, Tam2023Evaluating, Min2023FActScore} leverage external tools or neural networks (e.g., BERT-NLI, GPT-4) to score the faithfulness and factuality of the model output.
Recently, many works focus on suppressing this problem by retrieval-augmented generation (RAG)~\cite{Lewis2020Retrieval}, contrastive learning~\cite{Sun2023Contrastive}, contradictory evaluation~\cite{Mundler2023Self}, and decoding strategies~\cite{Lee2022Factuality, Shi2023Trusting, Li2023Inference}.
Different from them, we aim to solve the hallucination problem on graph tasks with preference alignment.

\section{Conclusion}

This paper proposes a novel {\model} framework that empowers the LLM with the capacity to solve graph reasoning and generation tasks.
To bridge the gap between graph data and textual language models, we introduce a structured format verbalizer to transform each graph into a code-like foemat and continually tune the LLM based on the instruction dataset, which is collected from 29 graph tasks. 
In addition, we also introduce a graph preference alignment stage to further mitigate the hallucination problem when reasoning on or generating a graph.
Extensive experiments illustrate that {\model} can unleash the LLMs' power of graph reasoning and generation, and substantially achieve the best performance.
In our future work, we aim to further improve the performance of our framework on both graph-centric and universal NLP tasks, and scale it to other LLMs.

\bibliography{anthology, custom}

\appendix

\begin{table*}[t]
\centering
\resizebox{\linewidth}{!}{
\begin{tabular}{cc cc ccc c}
\toprule
\multirow{2}*{\textbf{Clusters}} & \multirow{2}*{\textbf{Tasks}} & \multirow{2}*{\textbf{Source}} & \multirow{2}*{\textbf{Sampling}} & \multicolumn{2}{c}{\textbf{Instruction Dataset}} & \multicolumn{2}{c}{\textbf{Preference Dataset}}  \\
&  &  &  &\bf \textbf{\#Train} &\bf \textbf{\#Test} & \bf \textbf{\#Train} & \bf \textbf{\#Test} \\
\hline
\multirow{6}*{Structure} & Conn. Dect. & \cite{Wang2023Can} & Up & 3,737 & 237 & 2,227 & 463  \\
& Cycle Dect. & \cite{Wang2023Can} & Up & 2,877 & 191 & 863 & 191 \\
& Hami. Path & \cite{Wang2023Can} & Up & 1,315 & 55 & - & - \\
& Bipt. Match & \cite{Wang2023Can} & Up & 1,755 & 71 & - & - \\
& Shrt. Path & \cite{Wang2023Can} & Up & 1,580 & 64 & 948 & 128 \\
& Degree Comp. & \cite{Wang2023Can} & Up & 2,435 & 230 & 1,429 & 445 \\
\hline
\multirow{4}*{Caption} & Wikipedia & \cite{Wang2022Knowledge} & Down & 516,585 & 1,979 & 15,208 & 4,785  \\
& WebNLG & \cite{Gardent2017Creating} & $100\%$ & 12,237 & 2,000 & 6,040 & 2,616 \\
& GenWiki & \cite{Jin2020GenWiki} & $100\%$ & 99,997 & 1,000 & - & - \\
& EventNA & \cite{Colas2021EventNarrative} & $100\%$ & 58,733 & 1,952 & - & - \\
& Xalign & \cite{Abhishek2022XAlign} & $100\%$ & 30,000 & 470 & - & - \\
\hline
\multirow{4}*{Graph QA} & PathQSP & \cite{Zhou2018An} & Down & 30,530 & 1,000 & 27477 & 3,000  \\
& GrailQA & \cite{Gu2021Beyond} & Down & 13,797 & 1,421 & - & - \\
& WebQSP & \cite{Berant2013Semantic} & Down & 13,152 & 1,465 & - & - \\
& WikiTQ & \cite{Pasupat2015Compositional} & Down & 2,780 & 688 & - & - \\
\hline
\multirow{5}*{Node CLS} & Cora & \cite{McCallum2000Automating} & Down & 548 & 961 & 166 & 965  \\
& Citeseer & \cite{Giles1998CiteSeer} & Down & 943 & 995 & 284 & 990 \\
& Pubmed & \cite{Sen2008Collective} & Down & 9,736 & 1,756 & 2,988 & 1,789 \\
& Arxiv & \cite{Hu2020Open} & Down & 9,710 & 400 & 2,705 & 325 \\
& Products & \cite{Hu2020Open} & Down & 19,975 & 1,688 & 5,995 & 1,719 \\
\hline
\multirow{3}*{Link Pred.} & Wikidata & \cite{Wang2022Knowledge} & Down & 49,320 & 3,190 & - & -  \\
& FB15K-237 & \cite{Bollacker2008Freebase} & Down & 2,988 & 92 & - & - \\
& ConceptNet & \cite{Speer2017ConceptNet} & Down & 21,240 & 598 & - & - \\
\hline
Relevance & Wikipedia & \cite{Wang2022Knowledge} & Down & 39,672 & 1,991 & - & -  \\
\hline
\multirow{1}*{RecSys} & Amazon & \cite{He2016Ups} & Down & 2,424 & 250 & - & -  \\
\hline
\multirow{3}*{IE} & Wikipedia & \cite{Wang2022Knowledge} & Down & 73,101 & 1,814 & 19,490 & 1,589  \\
& UIE & \cite{Wang2023InstructUIE} & 100\% & 285,877 & 3,000 & - & - \\
& InstructKGC & \cite{Gui2023InstructIE} & Down & 31,605 & 994 & - & - \\
\hline
\multirow{1}*{Graph Gen.} & NLGraph & \cite{Wang2023Can} & Down & 3,056 & 407 & - & -  \\
\hline
\multicolumn{4}{l}{\textbf{The total number of the corpus}} & 1,341,885 & 30,959 & 85,820 & 19,005  \\
\bottomrule
\end{tabular}
}
\caption{The data statistics of each graph task for graph instruction tuning and preference alignment.}
\label{tab:instructgraph_corpus}
\end{table*}

\section{Details of the InstructGraph Corpus}
\label{app:corpus}

In this section, we provide some details of the corpus construction including both instruction and preference perspective.

\subsection{Instruction Tuning Dataset}
\label{app:instruction_corpus}
To merge all graph-oriented reasoning and generation tasks, we collect and construct 29 tasks to form instruction data.
We do not construct training sets for graph thought modeling.

\paragraph{Graph Structure Modeling}

Graph structure modeling aims to urge the LLM to understand the structure of a graph along with the corresponding task-specific instruction.
To reach this aim, we collect structure dataset NLGraph~\cite{Wang2023Can}.
The original dataset consists of 8 different tasks, such as \emph{Connectivity Detection}, \emph{Cycle Detection}, \emph{Topological Sorting}, S\emph{hortest Path Computing}, \emph{Maximum Flow Computing}, \emph{Bipartite Graph Matching}, \emph{Hamilton Path Detection} and \emph{GNN Embedding}.
Yet, the authors~\citet{Wang2023Can} mentioned that the current LLMs are hard to perform on more complex graph reasoning, such as \emph{Topological Sorting}, \emph{Maximum Flow Computing}, and \emph{GNN Embedding}, so we remove them.
In addition, we also random sample some graphs of NLGraph, and construct a \emph{Degree Computing} task.
\begin{itemize}
    \item Connectivity Detection: detect whether there exists a path between two nodes in the graph. This task is a binary classification and the answer should be 'The answer is yes' or 'The answer is no'.
    \item Cycle Detection: determine if there is a cycle in this graph. This task is a binary classification and the answer should be 'Yes' or 'No'.
    \item Topological Sorting: determine if there is a path that visits every node exactly once in this graph. This task is a binary classification and the answer should be 'Yes' or 'No'.
    \item Bipartite Graph Matching: detect whether there exists an edge between two given nodes in a bipartite graph. This task is a binary classification and the answer should be 'Yes' or 'No'.
    \item Shortest Path Computing: find the shortest path between two nodes in the graph, and calculate the sum of the weights in the shortest path. The answer is a sequence of the path with a value.
    \item Graph Degree Computing: calculate the degree of the target node in the graph. The answer is an integer value.
\end{itemize}

\begin{table*}[t]
\centering
\resizebox{\linewidth}{!}{
\begin{tabular}{c|l|l|l}
\toprule
\bf Task Name & \bf Hallucination Type &\bf Positive Answer &\bf Negative Answer \\
\midrule
\multirow{4}*{\makecell[l]{\\\\\\\textbf{Conn. Dect.}\\\textbf{Cycle Detect.}\\\textbf{Shrt. Path}\\\textbf{Degree Comp.}}} & \makecell[l]{Correct graph but wrong answer} & \makecell[l]{<The original answer>} & \makecell[l]{<Randomly sampled from other examples>} \\
\cline{2-4}
& \makecell[l]{Unfactual graph but wrong answer} & \makecell[l]{Sorry, the graph contains some wrong knowledge in the follow:\\ <list all unfactual triples>. So the question is unanswerable, \\you had better provide a correct graph.} & \makecell[l]{<The original answer>} \\
\cline{2-4}
& \makecell[l]{Conflict graph but wrong answer} & \makecell[l]{Sorry, the graph contains some conflict edges in the follow:\\ <list all conflict triples>. So the question is unanswerable, \\you had better provide a correct graph.} & \makecell[l]{<The original answer>} \\
\cline{2-4}
& \makecell[l]{Missing graph but wrong answer} & \makecell[l]{Sorry, the graph does not exist node {node name}.\\ So the question is unanswerable, you had better provide a \\correct graph.} & \makecell[l]{<The original answer>} \\
\midrule
\multirow{3}*{\makecell[l]{\\\\\\\\\textbf{Caption}}} & \makecell[l]{Correct graph but wrong answer} & \makecell[l]{<The original answer>} & \makecell[l]{<Randomly sampled from other examples>} \\
\cline{2-4}
& \makecell[l]{Unfactual graph but wrong answer} & \makecell[l]{Sorry, the graph contains some wrong knowledge in the follow:\\ <list all unfactual triples>. based on the corrected graph, \\the answer can be <The original answer>.} & \makecell[l]{<The original answer>} \\
\cline{2-4}
& \makecell[l]{Conflict graph but wrong answer} & \makecell[l]{Sorry, the graph contains some conflict edges in the follow:\\ <list all conflict triples>. So the question is unanswerable, \\you had better provide a correct graph.} & \makecell[l]{<The original answer>} \\

\midrule
\multirow{4}*{\makecell[l]{\\\\\\\\\textbf{Graph QA}}} & \makecell[l]{Correct graph but wrong answer} & \makecell[l]{<The original answer>} & \makecell[l]{<Randomly sampled from other examples>} \\
\cline{2-4}
& \makecell[l]{Unfactual graph but wrong answer} & \makecell[l]{Sorry, the graph contains some wrong knowledge in the follow:\\ <list all unfactual triples>. based on the corrected graph, \\the answer can be <The original answer>.} & \makecell[l]{<The original answer>} \\
\cline{2-4}
& \makecell[l]{Conflict graph but wrong answer} & \makecell[l]{Sorry, the graph contains some conflict edges in the follow:\\ <list all conflict triples>. So the question is unanswerable, \\you had better provide a correct graph.} & \makecell[l]{<The original answer>} \\
\cline{2-4}
& \makecell[l]{Missing graph but wrong answer} & \makecell[l]{Based on the world knowledge, the correct answer to the\\ question is <The original answer>, but the answer does not \\exist in the graph.} & \makecell[l]{<The original answer>} \\

\midrule
\multirow{1}*{\makecell[l]{\textbf{Node CLS}}} & \makecell[l]{Correct graph but wrong answer} & \makecell[l]{<The original answer>} & \makecell[l]{<Randomly sampled from other examples>} \\

\midrule
\multirow{4}*{\makecell[l]{\\\textbf{IE}}} & \makecell[l]{Wrong input but wrong graph} & \makecell[l]{<The original graph>} & \makecell[l]{<Randomly sampled from other examples>} \\
\cline{2-4}
& \makecell[l]{Correct input but unfaithful graph} & \makecell[l]{<The original graph>} & \makecell[l]{<Randomly edit entities in the original graph>} \\
\cline{2-4}
& \makecell[l]{Correct input but unfactual graph} & \makecell[l]{<Randomly edit edges in the original graph>} & \makecell[l]{<The original graph>} \\
\cline{2-4}
& \makecell[l]{Correct input but missing or \\redundant information in graph} & \makecell[l]{<Randomly remove or add edges in the original graph>} & \makecell[l]{<The original graph>} \\
\bottomrule
\end{tabular}
}
\caption{The positive and negative answer of each example for preference alignment.}
\label{tab:positive_and_negative}
\end{table*}

\paragraph{Graph Language Modeling}
Graph language modeling aims to teach the LLM to understand both the structure and semantics knowledge of the graph and answer the question.
We decompose this group into 6 kinds of tasks, including \emph{graph caption generation}, \emph{graph question answering}, \emph{graph node classification}, \emph{graph link prediction}, \emph{graph relevance inspection}, and \emph{graph collaboration filtering}.
\begin{itemize}
    \item Graph caption generation: generate an encyclopedia passage when given a knowledge graph with all entities and structure triples representing factual and commonsense knowledge. We directly choose the datasets from WebNLG~\cite{Gardent2017Creating}, GenWiki~\cite{Jin2020GenWiki}, EventNarrative~\cite{Colas2021EventNarrative}, XAlign~\cite{Abhishek2022XAlign}. In addition, we also follow~\cite{Wang2022Knowledge} to collect the Wikipedia corpus and corresponding wikidata knowledge graph to build the caption task. Specifically, we use the AC automatic machine algorithm to recognize all entities in the passage and construct a 2-hop sub-graph based on the topic entity.
    \item Graph question answering: find an entity and a reasoning path in the graph to answer the question. We directly collect the corpus from PathQuestions~\cite{Zhou2018An}, GrailQA~\cite{Gu2021Beyond}, WebQuestions~\cite{Berant2013Semantic}, WikiTableQuestions~\cite{Pasupat2015Compositional}. Especially, the WikiTableQuestions is a table understanding task that answers a question based on the table. To make our framework support this kind of task, we perform preprocessing that transforms each row line of the table into a single graph, where the table head is the relation name and each cell is the entity. 
    \item Graph node classification: classify the target node based on the corresponding graph. We directly choose from Cora~\cite{McCallum2000Automating}, Citeseer~\cite{Giles1998CiteSeer}, Pubmed~\cite{Sen2008Collective}, OGBN-ArXiv, and OGBN-Products~\cite{Hu2020Open}. Because the graph in these tasks is too big, we only sample a 2-hop sub-graph of centering each target node. We also perform down-sampling for each task.
    \item Graph link prediction: classify the edge (relation) between two given nodes (entities) based on the graph. We choose three main knowledge graph, such as Wikidata~\cite{Wang2021KEPLER}, Freebase~\cite{Bollacker2008Freebase}, ConceptNet~\cite{Speer2017ConceptNet}. Specifically, we random sample a subset of triples, and then extract and merge two 2-hop sub-graphs that center with two entities, respectively.
    \item Graph relevance inspection: inspect whether the caption is relevant to the graph. The task is a binary classification with two categories, i.e., "relevant" and "irrelevant". We directly use the same corpus from wikipedia~\cite{Wang2022Knowledge} in \emph{graph caption generation} task. For the negative sampling of each graph, we directly choose other captions.
    \item Graph Collaboration Filtering: predict the score that the user node prefers to the target item node based on the collaboration graph. We choose the widely used Amazon~\cite{He2016Ups} as the corpus. Because the Amazon dataset does not provide any graph data, we thus perform a preprocessing stage to construct a collaboration graph. Specifically, we calculate the Jaccard similarity between each pair of users based on their preference items and then recall the top-10 similarity users for each user to form a graph. Hence, we can inject this graph into the LLM to let it know how to recommend some items based on all potential users.
\end{itemize}

\paragraph{Graph Generation Modeling}
This group aims to guide the LLM to generate a graph in a code-like format. We consider two challenging graph generation domains, including, \emph{knowledge graph generation} and \emph{structure graph generation}.
\begin{itemize}
    \item Knowledge graph generation: similar to information extraction which aims to extract entities and relations when given one passage. We directly choose the corpus from unified information extraction (UIE)~\cite{Wang2023InstructUIE, Gui2023InstructIE}, which consists of 21 used named entity recognition (NER) tasks, 10 used relation extraction (RE), and 4 used event extraction (EE).
    \item Structure graph generation: generate a structure graph based on the description. For example, when given a graph description is ``Please generate a full-connection un-directed graph with four nodes ranging from 0 to 3.'', the expected code-like format graph is ``Graph[name='structure-graph']{node\_list=[0, 1, 2, 3]; edge\_list=[(0 <-> 1), (0 <-> 2), (0 <-> 3), (1 <-> 2), (1 <-> 3), (2 <-> 3)];}''. We can directly reuse the corpus from NLGraph~\cite{Wang2023Can} and sample a subset to build this task.
\end{itemize}

\subsection{Preference Alignment Dataset}
\label{app:preference_corpus}

We have selected a partial dataset from the graph instruction tuning dataset for preference alignment. This dataset includes Connection Detection, Cycle Detection, Shortest Path Computing, Degree Computing, Graph Caption with Wikipedia and WebNLG, Graph QA with PathQSP, Node CLS with Cora, Citeseer, Pubmed, Arxiv, and Products, and IE with Wikipedia.

For each task, we design positive and negative answers to support preference alignment. Details are shown in Table~\ref{tab:positive_and_negative}.

\end{document}